\documentclass[10pt,twocolumn,letterpaper]{article}

\usepackage{wacv}
\usepackage{fancyhdr}

\usepackage{lmodern}
\usepackage{graphicx}
\usepackage{multirow,tabularx}
\usepackage[point]{fltpoint}
\usepackage{textcomp}
\usepackage{amsmath,amssymb}
\usepackage{ifthen}
\usepackage{color}
\usepackage{xcolor}
\usepackage{tabularx}
\usepackage{booktabs}
\usepackage{longtable}
\usepackage{footmisc}

\usepackage[listofformat=parens, justification=centering, subrefformat=subparens]{subfig}  

\DeclareGraphicsExtensions{.pdf,.jpg,.png}
\graphicspath{{./graphics/}{./figures/}}

\newcolumntype{C}{X<{\centering}}

\definecolor{lightgreen}{rgb}{0.67, 0.88, 0.69}
\definecolor{darkgreen}{rgb}{0,0.55,0}
\definecolor{linkcolor}{rgb}{0,0,.65}

\newcommand\Caption[3][]{\caption[#2]{\label{#1}\textsc{#2}. \small#3}}
\renewcommand\etal[1]{\textit{et al.}~\cite{#1}}

\renewcommand\sec[1]{Sec.~\ref{sec:#1}}
\newcommand\fig[1]{Fig.~\ref{fig:#1}}

\newcommand\tab[1]{Tab.~\ref{tab:#1}}
\newcommand\stab[1]{Tab.~\subref{tab:#1}}

\DeclareMathOperator*{\argmax}{arg\,max}

\newcommand{\p}[1]{\ensuremath{P_{#1}}\xspace}

\newcommand\new[2][M]{#2}

\def\wacvPaperID{439} 

\wacvalgorithmstrack   

\wacvfinalcopy 

\ifwacvfinal
\usepackage[breaklinks=true,bookmarks=false]{hyperref}
\else
\usepackage[pagebackref=true,breaklinks=true,colorlinks,bookmarks=false]{hyperref}
\fi

\begin{document}

%
\title{Large-Scale Open-Set Classification Protocols for ImageNet}

\author{Andres Palechor \and Annesha Bhoumik \and Manuel G\"unther \and
Department of Informatics, University of Zurich, Andreasstrasse 15, CH-8050 Zurich \\
{\tt\small https://www.ifi.uzh.ch/en/aiml.html}
}

\maketitle
\thispagestyle{empty}

{
  \chead{\footnotesize This is a pre-print of the original paper accepted at the Winter Conference on Applications of Computer Vision (WACV) 2023.}
  \lhead{}
  \thispagestyle{fancy}
  \pagenumbering{gobble}
}


\begin{abstract}

Open-Set Classification (OSC) intends to adapt closed-set classification models to real-world scenarios, where the classifier must correctly label samples of known classes while rejecting previously unseen unknown samples.
Only recently, research started to investigate on algorithms that are able to handle these unknown samples correctly.
Some of these approaches address OSC by including into the training set negative samples that a classifier learns to reject, expecting that these data increase the robustness of the classifier on unknown classes.
Most of these approaches are evaluated on small-scale and low-resolution image datasets like MNIST, SVHN or CIFAR, which makes it difficult to assess their applicability to the real world, and to compare them among each other.
We propose three open-set protocols that provide rich datasets of natural images with different levels of similarity between known and unknown classes.
The protocols consist of subsets of ImageNet classes selected to provide training and testing data closer to real-world scenarios.
Additionally, we propose a new validation metric that can be employed to assess whether the training of deep learning models addresses both the classification of known samples and the rejection of unknown samples.
We use the protocols to compare the performance of two baseline open-set algorithms to the standard SoftMax baseline and find that the algorithms work well on negative samples that have been seen during training, and partially on out-of-distribution detection tasks, but drop performance in the presence of samples from previously unseen unknown classes.

\end{abstract}



\section{Introduction}

\begin{figure*}[t]
  \centering
	\includegraphics[width=.99\linewidth]{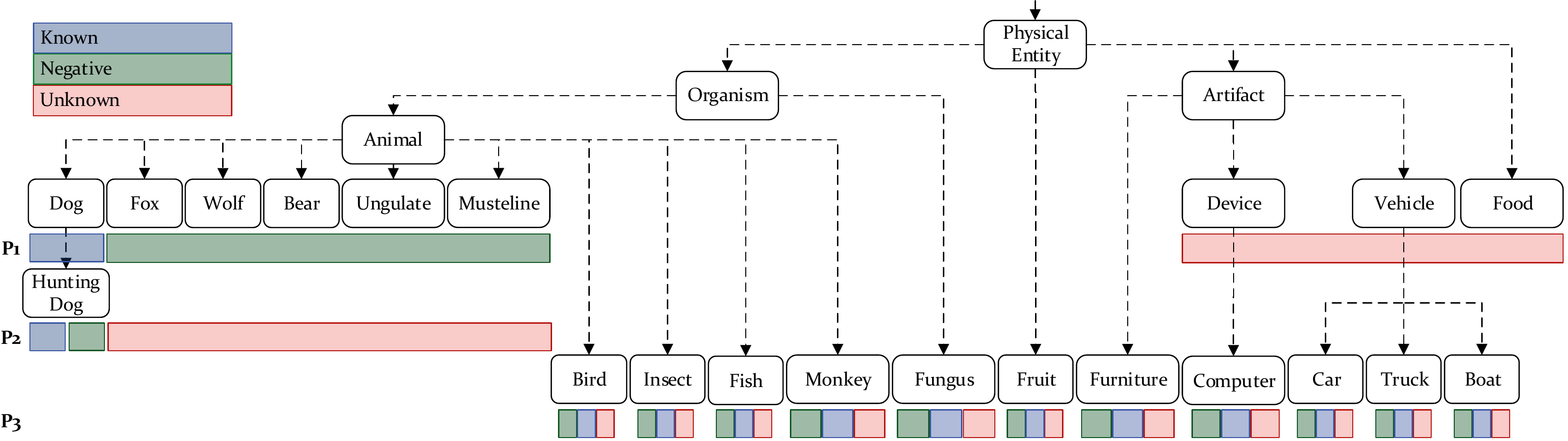}
	\Caption[fig:tree]{Class Sampling in our Open-Set Protocols}{
  We make use of the WordNet hierarchy \cite{miller1998wordnet} to define three protocols of different difficulties.
  In this figure, we show the superclasses from which we sample the final classes, all of which are leaf nodes taken from the ILSVRC 2012 dataset.
	Dashed lines indicate that the lower nodes are descendants, but they might not be direct children of the upper nodes.
  Additionally, all nodes have more descendants than those shown in the figure.
	The colored bars below a class indicate that its subclasses are sampled for the purposes shown in the top-left of the figure.
  For example, all subclasses of “Dog” are used as known classes in protocol \p1, while the subclasses of “Hunting Dog” are partitioned into known and negatives in protocol \p2.
	For protocol \p3, several intermediate nodes are partitioned into known, negative and unknown classes.
	}
\end{figure*}

Automatic classification of objects in images has been an active direction of research for several decades now.
The advent of Deep Learning has brought algorithms to a stage where they can handle large amounts of data and produce classification accuracies that were beyond imagination a decade before.
Supervised image classification algorithms have achieved tremendous success when it comes to detecting classes from a finite number of \textit{known} classes, what is commonly known as evaluation under the closed-set assumption.
For example, the deep learning algorithms that attempt the classification of ten handwritten digits \cite{lecun1998mnist} achieve more than 99\% accuracy when presented with a digit, but it ignores the fact that the classifier might be confronted with non-digit images during testing \cite{dhamija2018agnostophobia}.
Even the well-known ImageNet Large Scale Visual Recognition Challenge (ILSVRC) \cite{russakovsky2015imagenet} contains 1000 classes during training, and the test set contains samples from exactly these 1000 classes, while the real world contains many more classes, e.g., the WordNet hierarchy \cite{miller1998wordnet} currently knows more than 100'000 classes.\footnote{\url{https://wordnet.princeton.edu}}
Training a \new{categorical classifier} that can differentiate all these classes is currently not possible -- only feature comparison approaches \cite{radford2021learning} exist --  and, hence, we have to deal with samples that we do not know how to classify.

Only recently, research on methods to improve classification in presence of \emph{unknown} samples has gained more attraction.
These are samples from previously unseen classes that might occur during deployment of the algorithm in the real world and that the algorithm needs to handle correctly by not assigning them to any of the known classes.
Bendale and Boult \cite{bendale2016openmax} provided the first algorithm that incorporates the possibility to reject a sample as unknown into a deep network that was trained on a finite set of known classes.
Later, other algorithms were developed to improve the detection of unknown samples.
Many of these algorithms require to train on samples from some of the unknown classes that do not belong to the known classes of interest -- commonly, these classes are called \emph{known unknown} \cite{miller2018dropout}, but since this formulation is more confusing than \new{helpful}, we will term these classes as the \emph{negative} classes.
For example, Dhamija \etal{dhamija2018agnostophobia} employed samples from a different dataset, i.e., they trained their system on MNIST as known classes and selected EMNIST letters as negatives.
Other approaches try to create negative samples by utilizing known classes in different ways, e.g., Ge \etal{ge2017generative} used a generative model to form negative samples, while Zhou \etal{zhou2021placeholders} try to utilize internal representations of mixed known samples.

One issue that is inherent in all of these approaches \new{-- with only a few exceptions \cite{bendale2016openmax,rudd2017evm} --} is that they evaluate only on small-scale datasets with a few known classes, such as 10 classes in MNIST \cite{lecun1998mnist}, CIFAR-10 \cite{krizhevsky2009cifar}, SVHN\cite{netzer2011svhn} or mixtures of these.
While many algorithms claim that they can handle unknown classes, the number of known classes is low, and it is unclear whether these algorithms can handle more known classes, or more diverse sets of unknown classes.
\new{Only lately, a large-scale open-set validation protocol is defined on ImageNet \cite{vaze2022openset}, but it only separates unknown samples based on \emph{visual}\footnote{\new{In fact, Vaze \etal{vaze2022openset} do not specify their criteria to select unknown classes and only mention \emph{visual similarity} in their supplemental material.}} and not semantic similarity.}
Another issue of research on open-set classification is that most of the employed evaluation criteria, such as accuracy, macro-F1 or ROC metrics, do not evaluate open-set classification as it would be used in a real-world task.
Particularly, the currently employed validation metrics that are used during training a network do not reflect the target task and, thus, it is unclear whether the selected model is actually the best model for the desired task.

In this paper we, therefore, propose large-scale open-set recognition protocols that can be used to train and test various open-set algorithms -- and we will show-case the performance of three simple algorithms in this paper.
We decided to build our protocols based on the well-known and well-investigated ILSVRC 2012 dataset \cite{russakovsky2015imagenet}, and we build three evaluation protocols \p1, \p2 and \p3 that provide various difficulties based on the WordNet hierarchy \cite{miller1998wordnet}, as displayed in \fig{tree}.
The protocols are publicly available,\footnote{\label{fn:code}\url{https://github.com/AIML-IfI/openset-imagenet}} including  source code for the baseline implementations and the evaluation, which enables the reproduction of the results presented in this paper.
With these new protocols, we hope to foster more comparable and reproducible research into the direction of open-set object classification as well as related topics such as out-of-distribution detection.
This allows researchers to test their algorithms on our protocols and directly compare with our results.

The contributions of this paper are as follows:
\begin{itemize}
  \item We introduce three novel open-set evaluation protocols with different complexities for the ILSVRC 2012 dataset.
  \item We propose a novel evaluation metric that can be used for validation purposes when training open-set classifiers.
  \item We train deep networks with three different techniques and report their open-set performances.
  \item We provide all source code\footref{fn:code} for training and evaluation of our models to the research community.
\end{itemize}

\section{Related Work}
In open-set classification, a classifier is expected to correctly classify known test samples into their respective classes, and correctly detect that unknown test samples do not belong to any known class.
The study of unknown instances is not new in the literature.
For example, novelty detection, which is also known as anomaly detection and has a high overlap with out-of-distribution detection, focuses on identifying test instances that do not belong to training classes.
It can be seen as a binary classification problem that determines if an instance belongs to any of the training classes or not, but without exactly deciding which class \cite{bodesheim2015novelty}, and includes approaches in supervised, semi-supervised and unsupervised learning \cite{jiang2018trust, ren2019likelihood, golan2018anomaly}.

However, all these approaches only consider the classification of samples into known and unknown, leaving the later classification of known samples into their respective classes as a second step.
Ideally, these two steps should be incorporated into one method.
An easy approach would be to threshold on the maximum class probability of the SoftMax classifier using a confidence threshold, assuming that for an unknown input, the probability would be distributed across all the classes and, hence, would be low \cite{matan1990handwritten}.
Unfortunately, often inputs overlap significantly with known decision regions and tend to get misclassified as a known class with high confidence \cite{dhamija2018agnostophobia}.
It is therefore essential to devise techniques that are more effective than simply thresholding SoftMax probabilities in detecting unknown inputs.
Some initial approaches include extensions of 1-class and binary Support Vector Machines (SVMs) as implemented by Scheirer \etal{scheirer2013toward} and devising recognition systems to continuously learn new classes \cite{bendale2015openworld,rudd2017evm}.

While the above methods make use only of known samples in order to disassociate unknown samples, other approaches require samples of some negative classes, hoping that these samples generalize to all unseen classes.
For example, Dhamija \etal{dhamija2018agnostophobia} utilize negative samples to train the network to provide low confidence values for all known classes when presented with a sample from an unknown class.
Many researchers \cite{ge2017generative,yu2017opencategory,neal2018counterfactual} utilize generative adversarial networks to produce negative samples from the known samples.
Zhou \etal{zhou2021placeholders} combined pairs of known samples to define negatives, both in input space and deeper in the network.
Other approaches to open-set recognition are discussed by Geng \etal{geng2021recent}.

One problem that all the above methods possess is that they are evaluated on small-scale datasets with low-resolution images and low numbers of classes.
Such datasets include MNIST \cite{lecun1998mnist}, SVHN \cite{netzer2011svhn} and CIFAR-10 \cite{krizhevsky2009cifar} where oftentimes a few random classes are used as known and the remaining classes as unknown \cite{geng2021recent}.
Sometimes, other datasets serve the roles of unknowns, e.g., when MNIST build the known classes, EMNIST letters \cite{grother2016nist} are used as negatives and/or unknowns.
Similarly, the known classes are composed of CIFAR-10 and other classes from CIFAR-100 or SVHN are negatives or unknowns \cite{lakshminarayanan2017simple,dhamija2018agnostophobia}.
Only few papers make use of large-scale datasets such as ImageNet, where they either use the classes of ILSVRC 2012 as known and other classes from ImageNet as unknown \cite{bendale2016openmax,vaze2022openset}, or random partitions of ImageNet \cite{rudd2017evm,roady2020largescale}.

Oftentimes, evaluation protocols are home-grown and, thus, the comparison across algorithms is very difficult.
Additionally, there is no clear distinction on the similarities between known, negative and unknown classes, which makes it impossible to judge in which scenarios a method will work, and in which not.
Finally, the employed evaluation metrics are most often not designed for open-set classification and, hence, fail to address typical use-cases of open-set recognition.

\section{Approach}

\subsection{ImageNet Protocols}

\begin{table*}[t!]
  \Caption[tab:Protocolsb]{ImageNet Classes used in the Protocols}{
    This table shows the ImageNet parent classes that were used to create the three protocols.
    Known and negative classes are used for training the open-set algorithms, while known, negative and unknown classes are used in testing.
    \new{Given are the numbers of \emph{classes: training / validation / test} samples.}
    }
    \centering
\small
\renewcommand{\arraystretch}{1.3}
\begin{tabular}{p{0.015\textwidth} p{0.28\textwidth}<\centering p{0.28\textwidth}<\centering p{0.28\textwidth}<\centering}
\toprule
    & Known        & Negative       & Unknown       \\
    \midrule
\multirow{2}{*}{\p1}
    & All dog classes             & Other 4-legged animal classes& Non-animal classes\\[-.5ex]
    & 116: 116218 / 29055 / 5800 &  67: 69680 / 17420 / 3350 & 166 : --- /  ---  / 8300\\[.5ex]
\multirow{2}{*}{\p2}
    & Half of hunting dog classes & Half of hunting dog classes & Other 4-legged animal classes\\[-.5ex]
    & 30: 28895 / 7224 / 1500   &  31:  31794 / 7949 / 1550 &  55:  ---  /  ---  / 2750\\[.5ex]
\multirow{3}{*}{\p3}
    & Mix of common classes including animals, plants and objects
    & Mix of common classes including animals, plants and objects
    & Mix of common classes including animals, plants and objects\\[-.5ex]
    & 151: 154522 / 38633 / 7550 & 97: 98202 / 24549 / 4850  &	164: --- / --- / 8200 \\
\bottomrule
\end{tabular}
\end{table*}

Based on \cite{bhoumik2021master}, we design three different protocols to create three different artificial open spaces, with increasing level of similarity in appearance between inputs -- and increasing complexity and overlap between features -- of known and unknown classes.
To allow for the comparison of algorithms that require negative samples for training, we carefully design and include negative classes into our protocols.
This also allows us to compare how well these algorithms work on previously seen negative classes and how on previously unseen unknown classes.

In order to define our three protocols, we make use of the WordNet hierarchy that provides us with a tree structure for the 1000 classes of ILSVRC 2012.
Particularly, we exploit the \texttt{robustness} Python library \cite{engstrom2019robustness} to parse the ILSVRC tree.
All the classes in ILSVRC are represented as \new{leaf} nodes of that graph, and we use descendants of several intermediate nodes to form our known and unknown classes.
The definition of the protocols and their open-set partitions are presented in \fig{tree}, a more detailed listing of classes can be found in the supplemental material.
We design the protocols such that the difficulty levels of closed- and open-set evaluation varies.
While protocol \p1 is easy for open-set, it is hard for closed-set classification.
On the contrary, \p3 is more easy for closed-set classification and more difficult in open-set.
Finally, \p2 is somewhere in the middle, but small enough to run hyperparameter optimization that can be transferred to \p1 and \p3.

In the first protocol \p1, known and unknown classes are semantically quite distant, and also do not share too many visual features.
We include all 116 dog classes as known classes \new{-- since dogs represent the largest fine-grained intermediate category in ImageNet which makes closed-set classification difficult --} and select 166 non-animal classes as unknowns.
\p1 can, therefore, be used to test out-of-distribution detection algorithms since knowns and unknowns are not very similar.
In the second protocol \p2, we only look into the animal classes.
Particularly, we use several hunting dog classes as known and other classes of 4-legged animals as unknown.
This means that known and unknown classes are still semantically relatively distant, but image features such as fur is shared between known and unknown.
This will make it harder for out-of-distribution detection algorithms to perform well.
Finally, the third protocol \p3 includes ancestors of various different classes, both as known and unknown classes, by making use of the \texttt{mixed\_13} classes defined in the \texttt{robustness} library.
Since known and unknown classes come from the same ancestors, it is very unlikely that out-of-distribution detection algorithms will be able to discriminate between them, and real open-set classification methods need to be applied.

To enable algorithms that require negative samples, the negative classes are selected semantically similar to the known or at least in-between the known and the unknown.
It has been shown that selecting negative samples too far from the known classes does not help in creating better-suited open-set algorithms \cite{dhamija2018agnostophobia}.
Naturally, we can only define semantic similarity based on the WordNet hierarchy, but it is unclear whether these negative classes are also structurally similar to the known classes.
\tab{Protocolsb} displays a summary of the parent classes used in the protocols, and a detailed list of all classes is presented in the supplemental material.

Finally, we split our data into three partitions, one for training, one for validation and one for testing.
The training and validation partitions are taken from the original ILSVRC 2012 training images by randomly splitting off 80\% for training and 20\% for validation.
Since training and validation partitions are composed of known and negative data only, no unknown data is provided here.
The test partition is composed of the original ILSVRC validation set containing 50 images per class, and is available for all three groups of data: known, negative and unknown.
This assures that during testing no single image is used that the network has seen in any stage of the training.

\subsection{Open-Set Classification Algorithms}
We select three different techniques to train deep networks.
While other algorithms shall be tested in future work, we rely on three simple, very similar and well-known methods.
In particular, all three loss \new{functions} solely utilize the plain categorical cross-entropy loss $\mathcal J_{\mathrm{CCE}}$ on top of SoftMax activations (often termed as the SoftMax loss) in different settings.
Generally, the weighted categorical cross-entropy loss is:\\[-1ex]
\begin{equation}
  \label{eq:cce}
  \mathcal J_{\mathrm{CCE}} = \new{-}\frac1N \sum\limits_{n=1}^N \sum\limits_{c=1}^C w_c t_{n,c} \log y_{n,c}
\end{equation}
where $N$ is the number of samples in our dataset (note that we utilize batch processing), $t_{n,c}$ is the target label of the $n$th sample for class $c$, $w_c$ is a class-weight for class $c$ and $y_{n,c}$ is the output probability of class $c$ for sample $n$ using SoftMax activation:\\[-1ex]
\begin{equation}
  \label{eq:softmax}
  y_{c,n} = \frac{e^{z_{c,n}}}{\sum\limits_{c'=1}^C e^{z_{c',n}}}
\end{equation}
of the logits $z_{n,c}$, which are the network outputs.

The three different training approaches differ with respect to the targets $t_{n,c}$ and the weights $w_c$, and how negative samples are handled.
The first approach is the plain SoftMax loss (S) that is trained on only samples from the $K$ known classes.
In this case, the number of classes $C=K$ is equal to the number of known classes, and the targets are computed as one-hot encodings:
\begin{equation}
  \label{eq:targets-one-hot}
  \forall n, c\in\{1,\ldots,C\}: t_{n,c} = \begin{cases}1 & c = \tau_n \\ 0 & \text{otherwise} \end{cases}
\end{equation}
where $1\leq\tau_n\leq K$ is the label of the sample $n$.
For simplicity, we select the weights for each class to be identical: $\forall c: w_c = 1$, which is the default behavior when training deep learning models on ImageNet.
By thresholding the maximum probability $\max\limits_c y_{c,n}$, cf.~\sec{evaluation}, this method can be turned into a simple out-of-distribution detection algorithm.

The second approach is often found in object detection models \cite{dhamija2020elephant} which collect a lot of negative samples from the background of the training images.
Similarly, this approach is used in other methods for open-set learning, such as G-OpenMax \cite{ge2017generative} or PROSER \cite{zhou2021placeholders}.\footnote{While these methods try to sample better negatives for training, they rely on this additional class for unknown samples.}
In this Background (BG) approach, the negative data is added as an additional class, so that we have a total of $C=K+1$ classes.
Since the number of negative samples is usually higher than the number for known classes, we use class weights to balance them:
\begin{equation}
  \label{eq:weights-bg}
  \forall c\in\{1,...,C\}: w_c = \frac{N}{CN_c}
\end{equation}
where $N_c$ is the number of training samples for class $c$.
Finally, we use one-hot encoded targets $t_{n,c}$ according to \eqref{eq:targets-one-hot}, including label $\tau_n=K+1$ for negative samples.

As the third method, we employ the Entropic Open-Set (EOS) loss \cite{dhamija2018agnostophobia}, which is a simple extension of the SoftMax loss.
Similar to our first approach, we have one output for each of the known classes: $C=K$.
For known samples, we employ one-hot-encoded target values according to \eqref{eq:targets-one-hot}, whereas for negative samples we use identical target values:\\[-2ex]
\begin{equation}
  \label{eq:targets-equal}
  \forall n, c\in\{1,\ldots,C\}: t_{n,c} = \frac1C
\end{equation}
Sticking to the implementation of Dhamija \etal{dhamija2018agnostophobia}, we select the class weights to be $\forall c\colon w_c = 1$ for all classes including the negative class, and leave the optimization of these values for future research.

\subsection{Evaluation Metrics}
\label{sec:evaluation}
Evaluation of open-set classification methods is a more tricky business.
First, we must differentiate between validation metrics to monitor the training process and testing methods for the final reporting.
Second, we need to incorporate both types of algorithms, the ones that provide a separate probability for the unknown class and those that do not.

The final evaluation on the test set should differentiate between the behavior of known and unknown classes, and at the same time include the accuracy of the known classes.
Many evaluation techniques proposed in the literature do not follow these requirements.
For example, computing the area under the ROC curve (AUROC) will only consider the binary classification task: known or unknown, but does not tell us how well the classifier performs on the known classes.
Another metric that is often applied is the macro-F1 metric \cite{bendale2016openmax} that balances precision and recall for a K+1-fold binary classification task.
This metric has many properties that are counter-intuitive in the open-set classification task.
First, a different threshold is computed for each of the classes, so it is possible that the same sample can be classified both as one or more known classes and as unknown.
These thresholds are even optimized on the test set, and often only the maximum F1-value is reported.
Second, the method requires to define a particular probability of being unknown, which is not provided by two of our three networks.
Finally, the metric does not distinguish between known and unknown classes, but just treats all classes identically, but consequences of classifying an unknown sample as known are different from misclassifying a known sample.

The evaluation metric that follows our intuition best is the Open-Set Classification Rate (OSCR), which handles known and unknown samples separately \cite{dhamija2018agnostophobia}.
Based on a single probability threshold $\theta$, we compute both the Correct Classification Rate (CCR) and the False Positive Rate (FPR):\\[-5ex]

{\small
\begin{align}
  \label{eq:ccr-fpr}
  \mathrm{CCR}(\theta) &= \frac{\bigl|\{x_n \mid \tau_n \leq K \wedge \argmax\limits_{1\leq c\leq K} y_{n,c} = \tau_n \wedge y_{n,c} > \theta \}\bigr|}{|N_K|} \nonumber\\
  \mathrm{FPR}(\theta) &= \frac{\bigl|\{x_n \mid \tau_n > K \wedge \max\limits_{1\leq c\leq K} y_{n,c} > \theta \}\bigr|}{|N_U|}
\end{align}
}%
where $N_K$ and $N_U$ are the total numbers of known and unknown test samples, while $\tau_n \leq K$ indicates a known sample and $\tau_n > K$ refers to an unknown test sample.
By varying the threshold $\theta$ between 0 and 1, we can plot the OSCR curve \cite{dhamija2018agnostophobia}.
A closer look to \eqref{eq:ccr-fpr} reveals that the maximum is only taken over the known classes, purposefully leaving out the probability of the unknown class in the BG approach.\footnote{\new{A low probability for the unknown class does not indicate a high probability for any of the known classes. Therefore, the unknown class probability does not add any useful information.}}
Finally, this definition differs from \cite{dhamija2018agnostophobia} in that we use a $>$ sign for both FPR and CCR when comparing to $\theta$, which is critical when SoftMax probabilities of unknown samples reach 1 to the numerical limits.

Note that the computation of the Correct Classification Rate -- which is highly related to the classification accuracy on the known classes -- has the issue that it might be biased when known classes have different amount of samples.
Since the number of test samples in our known classes is always balanced, in our evaluation we are not affected by this bias, so we leave the adaptation of that metric to unbalanced datasets as future work.
Furthermore, the metric just averages over all \new{samples}, telling us nothing about different behavior of different classes -- it might be possible that one known class is always classified correctly while another class never is.
For a better inspection of these cases, open-set adaptations to confidence matrices need to be developed in the future.

\subsection{Validation Metrics}
For validation on SoftMax-based systems, often classification accuracy is used as the metric.
In open-set classification, this is not sufficient since we need to balance between accuracy on the known classes and on the negative class.
While using (weighted) accuracy might work well for the BG approach, networks trained with standard SoftMax and EOS do not provide a probability for the unknown class and, hence, accuracy cannot be applied for validation here.
Instead, we want to make use of the SoftMax scores to evaluate our system.

Since the final goal is to find a threshold $\theta$ such that the known samples are differentiated from unknown samples, we propose to compute the validation metric using our confidence metric:
\begin{equation}
  \label{eq:confidence}
  \begin{gathered}
    \gamma^- = \frac1{N_N} \sum\limits_{n=1}^{N_N} \biggl(1 - \max\limits_{1\leq c\leq K} y_{n,c} + \delta_{C,K} \frac1K \biggr)\\
    \gamma^+ = \frac1{N_K} \sum\limits_{n=1}^{N_K} y_{\tau_n} \qquad \gamma = \frac{\gamma^+ + \gamma^-}2
  \end{gathered}
\end{equation}
For known samples, $\gamma^+$ simply averages the SoftMax score for the correct class, while for negative samples, $\gamma^-$ computes the average deviation from the minimum possible SoftMax score, which is 0 in case of the BG class (where $C=K+1$), and $\frac1K$ if no additional background class is available ($C=K$).
When summing over all known and negative samples, we can see how well our classifier has learned to separate known from negative samples.
The maximum $\gamma$ score is 1 when all known samples are classified as the correct class with probability 1, and all negative samples are classified as any known class with probability 0 or $\frac1K$.
When looking at $\gamma^+$ and $\gamma^-$ individually, we can also detect if the training focuses on one part more than on the other.

\begin{figure*}[t!]
	\begin{center}
	\includegraphics[width=0.99\linewidth,page=1]{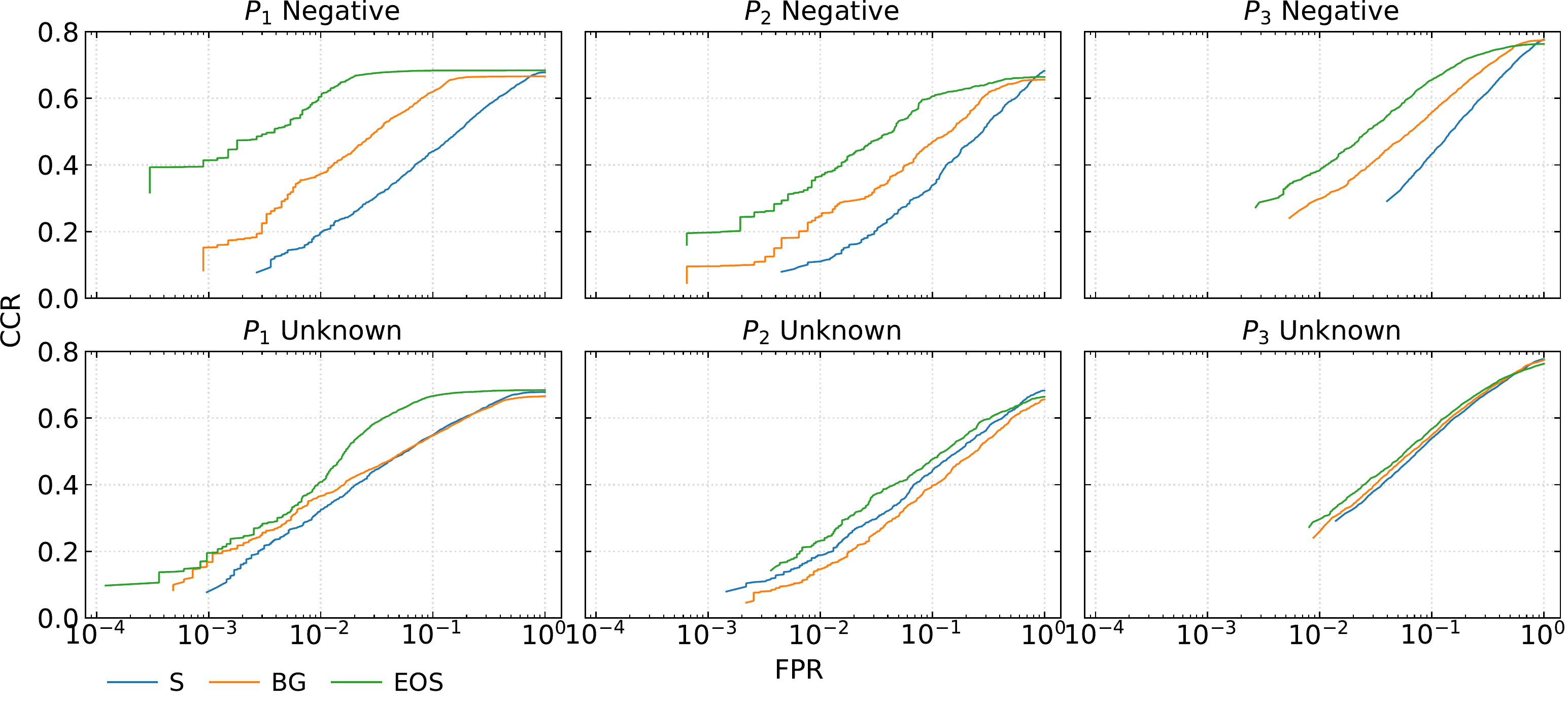}
	\end{center}
	\Caption[fig:oscr_all]{Open-Set Classification Rate Curves}{
	\new{OSCR curves are shown for test data of each protocol.
	The top row is calculated using negative test samples, while the bottom row uses unknown test samples.}
	Curves that do not extend to low FPR values indicate that the threshold in \eqref{eq:ccr-fpr} is maximized at $\theta=1$.
}
\end{figure*}

\section{Experiments}

Considering that our goal is not to achieve the highest closed-set accuracy but to analyze the performance of open-set algorithms in our protocols, we use a ResNet-50 model \cite{he2015deep} as it achieves low classification errors on ImageNet, trains rapidly, and is commonly used in the image classification task.
We add one fully-connected layer with $C$ nodes.
For each protocol, we train models using the three loss functions SoftMax (S), SoftMax with Background class (BG) and Entropic Open-Set (EOS) loss.
Each network is trained for 120 epochs using Adam optimizer with a learning rate of $10^{-3}$ and default beta values of 0.9 and 0.999.
Additionally, we use standard data preprocessing, i.e., first, the smaller dimension of the training images is resized to 256 pixels, and then a random crop of 224 pixels is selected.
Finally, we augment the data using a random horizontal flip with a probability of 0.5.

\fig{oscr_all} shows OSCR curves for the three methods on our three protocols using logarithmic FPR axes -- for linear FPR axes, please refer to the supplemental material.
We plot \new{the test set performance for both the negative and the unknown test samples}.
Hence, we can see how the methods work with unknown samples of classes\footnote{Remember that the known and negative sets are split into training and test samples so that we never evaluate with samples seen during training.} that have or have not been seen during training.
\new{We can observe} that all three classifiers in every protocol reach similar CCR values in the closed-set case (FPR=1), in some cases, EOS or BG even outperform the baseline.
This is good news since, oftentimes, open-set classifiers trade their open-set capabilities for reduced closed-set accuracy.
\new{In the supplemental material, we also provide a table with detailed} CCR values for particular selected FPRs, and $\gamma^+$ and $\gamma^-$ values computed on the test set.

Regarding the performance on negative samples of the test set, we can see that BG and EOS outperform the SoftMax (S) baseline, indicating that the classifiers learn to discard negatives.
Generally, EOS seems to be better than BG in this task.
Particularly, in \p1 EOS reaches a high CCR at FPR=$10^{-2}$, showing the classifier can easily reject the negative samples, which is to be expected since negative samples are semantically and structurally far from the known classes.

When evaluating the unknown samples of the test set that belong to classes that have not been seen during training, BG and EOS classifiers drop performance, and \new{compared to the gains on the validation set} the behavior is almost similar to the SoftMax baseline.
Especially when looking into \p2 and \p3, training with the negative samples does not clearly improve the open-set classifiers.
However, in \p1 EOS still outperforms S and BG for higher FPR, indicating the classifier learned to discard unknown samples up to some degree.
This shows that the easy task of rejecting samples very far from the known classes can benefit from EOS training with negative samples, i.e., the denoted open-set method is good for out-of-distribution detection, but not for the more general task of open-set classification.

\section{Discussion}

After we have seen that the methods perform well on negative and not so well on unknown data, let us analyze the results.
First, we show how our novel validation metrics can be used to identify gaps and inconsistencies during training of the open-set classifiers BG and EOS.
\fig{confidence} shows the confidence progress across the training epochs.
During the first epochs, the confidence of the known samples ($\gamma^+$, left in \fig{confidence}) is low since the SoftMax activations produce low values for all classes.
As the training progresses, the models learn to classify known samples, increasing the correct SoftMax activation of the target class.
Similarly, because of low activation values, the confidence of negative samples ($\gamma^-$, right) is close to 1 at the beginning of the training.
Note that EOS keeps the low activations during training, learning to respond only to known classes, particularly in \p1, where values are close to 1 during all epochs.
On the other hand, BG provides lower confidences for negative samples ($\gamma^-$).
This indicates that the class balancing technique in \eqref{eq:weights-bg} might have been too drastic and that higher weights for negative samples might improve results of the BG method.
Similarly, employing lower weights for the EOS classifier might improve the confidence scores for known samples at the cost of lower confidence for negatives.
Finally, from an open-set perspective, our confidence metric provides insightful information about the model training; so far, we have used it to explain the model performance, but together with more parameter tuning, the joint $\gamma$ metric can be used as criterion for early stopping as shown in the supplemental material.

We also analyze the SoftMax scores according to \eqref{eq:softmax} of known and unknown classes in every protocol.
For samples from known classes we use the SoftMax score of the correct class, while for unknown samples we take the maximum SoftMax score of any known class.
This enables us to make a direct comparison between different approaches in \fig{histogram}.
When looking at the score distributions of the known samples, we can see that many samples are correctly classified with a high probability, while numerous samples provide almost 0 probability for the correct class.
This indicates that a more detailed analysis, possibly via a confusion matrix, is required to further look into the details of these errors, but this goes beyond the aim of this paper.

\begin{figure}[t!]
	\begin{center}
	\includegraphics[width=.93\linewidth,page=2]{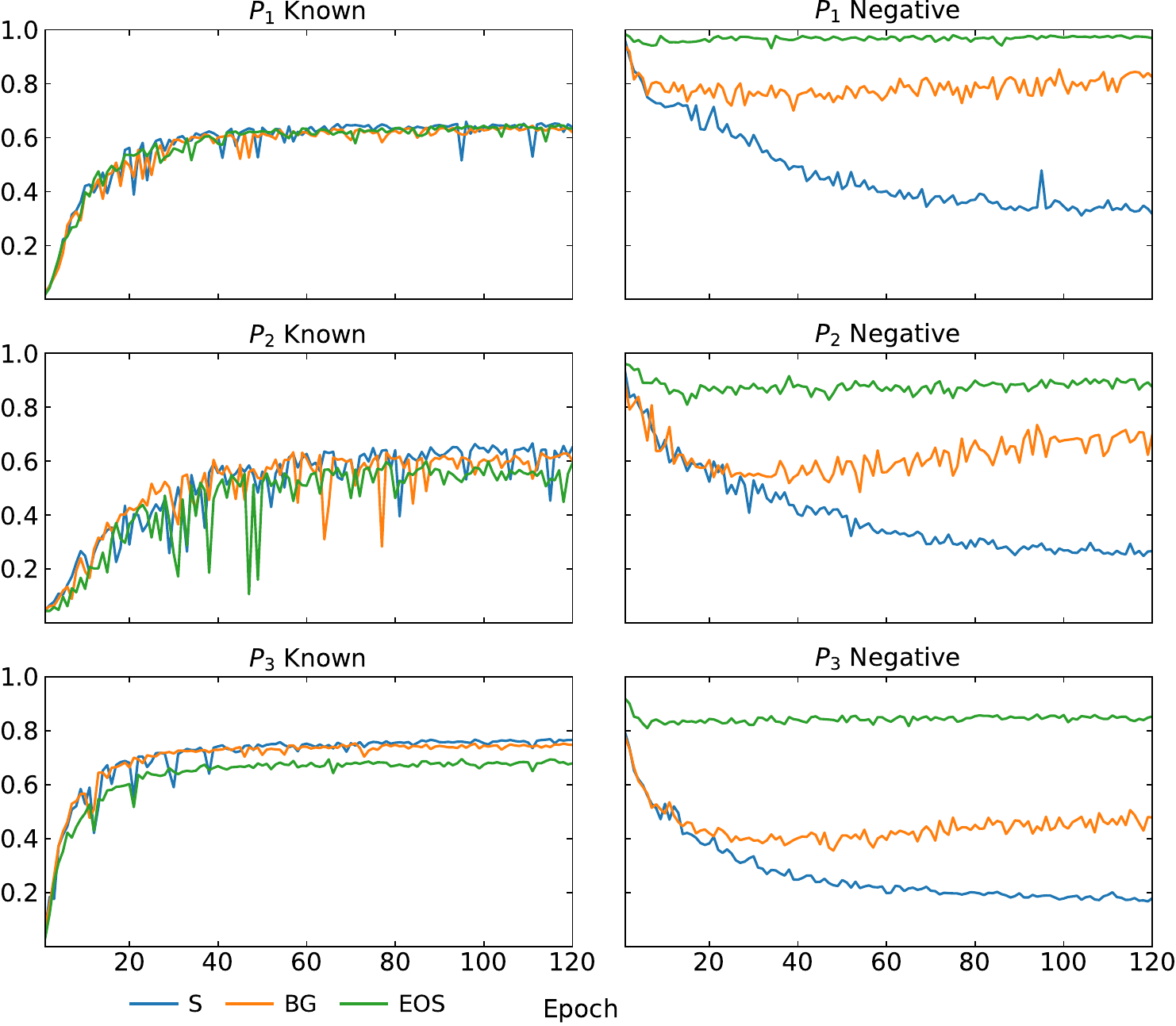}
	\end{center}
	\Caption[fig:confidence]{Confidence Propagation}{
	Confidence values according to \eqref{eq:confidence} are shown across training epochs of S, BG and EOS classifiers.
	On the left, we show the confidence of the known samples ($\gamma^+$), while on the right the confidence of negative samples ($\gamma^-$) is displayed.
		}
\end{figure}

More interestingly, the distribution of scores for unknown classes differ dramatically between approaches and protocols.
For \p1, EOS is able to suppress high scores almost completely, whereas both S and BG still have the majority of the cases providing high probabilities of belonging to a known class.
For \p2 and, particularly, \p3 a lot of unknown samples get classified as a known class with very high probability, throughout the evaluated methods.
Interestingly, the plain SoftMax (S) method has relatively high probability scores for unknown samples, especially in \p2 and \p3 where known and unknown classes are semantically similar.

\begin{figure}[t!]
	\begin{center}
	\includegraphics[width=.99\linewidth,page=3]{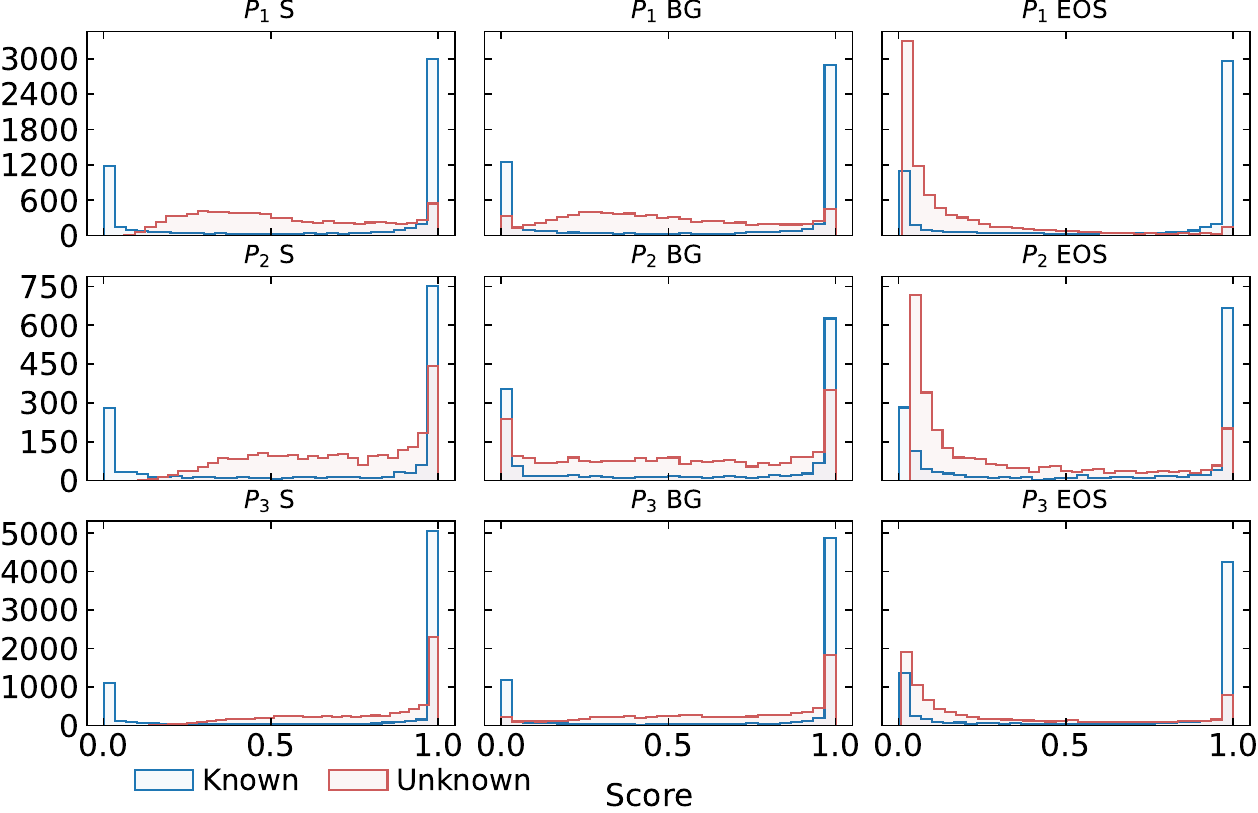}
	\end{center}
	\Caption[fig:histogram]{Histograms of SoftMax Scores}{
		\new{We evaluate SoftMax probability scores for all three methods and all three protocols.}
    For known samples, we present histograms of SoftMax score of the target class.
    For unknown samples, we plot the maximum SoftMax score of any known class.
    For S and EOS, the minimum possible value of the latter is $\frac1K$, which explains the gaps on the left-hand side.
	}
\end{figure}

\section{Conclusion}

In this work, we propose three novel evaluation protocols for open-set image classification that rely on the ILSVRC 2012 dataset and allow an extensive evaluation of open-set algorithms.
The data is entirely composed of natural images and designed to have various levels of similarities between its partitions.
Additionally, we carefully select the WordNet parent classes that allow us to include a larger number of known, negative and unknown classes.
In contrast to previous work, the class partitions are carefully designed, and we move away from implementing mixes of several datasets (where rejecting unknown samples could be relatively easy) and the random selection of known and unknown classes inside a dataset.
This allows us to differentiate between methods that work well in out-of-distribution detection, and those that really perform open-set classification.
A more detailed comparison of the protocols is provided in the supplemental material.

We evaluate the performance of three classifiers in every protocol using OSCR curves and our proposed confidence validation metric.
Our experiments show that the two open-set algorithms can reject negative samples, where samples of the same classes have been seen during training, but face a performance degradation in the presence of unknown data from previously unseen classes.
For a more straightforward scenario such as \p1, it is advantageous to use negative samples during EOS training.
While this result agrees with \cite{dhamija2018agnostophobia}, the performance of BG and EOS in \p2 and \p3 shows that these methods are not ready to be employed in the real world, and more parameter tuning is required to improve performances.
Furthermore, making better use of or augmenting the negative classes also poses a challenge in further research in open-set methods.

Providing different conclusions for the three different protocols reflects the need for evaluation methods in scenarios designed with different difficulty levels, which we provide within this paper.
Looking ahead, with the novel open-set classification protocols on ImageNet we aim to establish comparable and standard evaluation methods for open-set algorithms in scenarios closer to the real world.
Instead of using low-resolution images and randomly selected samples from CIFAR, MNIST or SVHN, we expect that our open-set protocols will establish benchmarks and promote reproducible research in open-set classification.
In future work, we will investigate and optimize more different open-set algorithms and report their performances on our protocols.

{\small
\bibliographystyle{ieee_fullname}
\bibliography{MG-References,MG-Publications}

\begin{thebibliography}{10}\itemsep=-1pt

\bibitem{bendale2015openworld}
Abhijit Bendale and Terrance Boult.
\newblock Towards open world recognition.
\newblock In {\em Conference on Computer Vision and Pattern Recognition
  (CVPR)}. IEEE, 2015.

\bibitem{bendale2016openmax}
Abhijit Bendale and Terrance~E. Boult.
\newblock Towards open set deep networks.
\newblock In {\em Conference on Computer Vision and Pattern Recognition
  (CVPR)}. IEEE, 2016.

\bibitem{bhoumik2021master}
Annesha Bhoumik.
\newblock Open-set classification on {ImageNet}.
\newblock Master's thesis, University of Zurich, 2021.

\bibitem{bodesheim2015novelty}
Paul Bodesheim, Alexander Freytag, Erik Rodner, and Joachim Denzler.
\newblock Local novelty detection in multi-class recognition problems.
\newblock In {\em Winter Conference on Applications of Computer Vision (WACV)},
  2015.

\bibitem{dhamija2020elephant}
Akshay Dhamija, Manuel G\"unther, Jonathan Ventura, and Terrance~E. Boult.
\newblock The overlooked elephant of object detection: Open set.
\newblock In {\em Winter Conference on Applications of Computer Vision (WACV)},
  2020.

\bibitem{dhamija2018agnostophobia}
Akshay~Raj Dhamija, Manuel G{\"u}nther, and Terrance~E. Boult.
\newblock Reducing network agnostophobia.
\newblock In {\em Advances in Neural Information Processing Systems (NeurIPS)},
  2018.

\bibitem{engstrom2019robustness}
Logan Engstrom, Andrew Ilyas, Shibani Santurkar, and Dimitris Tsipras.
\newblock Robustness (python library), 2019.

\bibitem{ge2017generative}
Zongyuan Ge, Sergey Demyanov, and Rahil Garnavi.
\newblock Generative {OpenMax} for multi-class open set classification.
\newblock In {\em British Machine Vision Conference (BMVC)}, 2017.

\bibitem{geng2021recent}
Chuanxing Geng, Sheng-Jun Huang, and Songcan Chen.
\newblock Recent advances in open set recognition: A survey.
\newblock {\em Transactions on Pattern Analysis and Machine Intelligence
  (TPAMI)}, 43(10):3614--3631, 2021.

\bibitem{golan2018anomaly}
Izhak Golan and Ran El-Yaniv.
\newblock Deep anomaly detection using geometric transformations.
\newblock In {\em Advances in Neural Information Processing Systems (NeurIPS)},
  2018.

\bibitem{grother2016nist}
Patrick Grother and Kayee Hanaoka.
\newblock {NIST} special database 19 handprinted forms and characters 2nd
  edition.
\newblock Technical report, National Institute of Standards and Technology
  (NIST), 2016.

\bibitem{he2015deep}
Kaiming He, Xiangyu Zhang, Shaoqing Ren, and Jian Sun.
\newblock Deep residual learning for image recognition.
\newblock In {\em Conference on Computer Vision and Pattern Recognition
  (CVPR)}, 2016.

\bibitem{jiang2018trust}
Heinrich Jiang, Been Kim, Melody Guan, and Maya Gupta.
\newblock To trust or not to trust a classifier.
\newblock In {\em Advances in Neural Information Processing Systems (NeurIPS)},
  2018.

\bibitem{krizhevsky2009cifar}
Alex Krizhevsky and Geoffrey Hinton.
\newblock Learning multiple layers of features from tiny images.
\newblock Technical report, University of Toronto, 2009.

\bibitem{lakshminarayanan2017simple}
Balaji Lakshminarayanan, Alexander Pritzel, and Charles Blundell.
\newblock Simple and scalable predictive uncertainty estimation using deep
  ensembles.
\newblock In {\em Advances in Neural Information Processing Systems (NIPS)},
  2017.

\bibitem{lecun1998mnist}
Yann LeCun, Corinna Cortes, and Christopher J.~C. Burges.
\newblock The {MNIST} database of handwritten digits, 1998.

\bibitem{matan1990handwritten}
Ofer Matan, R.K. Kiang, C.E. Stenard, B. Boser, J.S. Denker, D. Henderson, R.E.
  Howard, W. Hubbard, L.D. Jackel, and Yann Le~Cun.
\newblock Handwritten character recognition using neural network architectures.
\newblock In {\em USPS Advanced Technology Conference}, 1990.

\bibitem{miller2018dropout}
Dimity Miller, Lachlan Nicholson, Feras Dayoub, and Niko S{\"u}nderhauf.
\newblock Dropout sampling for robust object detection in open-set conditions.
\newblock In {\em International Conference on Robotics and Automation (ICRA)}.
  IEEE, 2018.

\bibitem{miller1998wordnet}
George~A Miller.
\newblock {\em WordNet: An electronic lexical database}.
\newblock MIT press, 1998.

\bibitem{neal2018counterfactual}
Lawrence Neal, Matthew Olson, Xiaoli Fern, Weng-Keen Wong, and Fuxin Li.
\newblock Open set learning with counterfactual images.
\newblock In {\em European Conference on Computer Vision (ECCV)}, 2018.

\bibitem{netzer2011svhn}
Yuval Netzer, Tao Wang, Adam Coates, Alessandro Bissacco, Bo Wu, and Andrew~Y.
  Ng.
\newblock Reading digits in natural images with unsupervised feature learning.
\newblock In {\em Advances in Neural Information Processing Systems (NIPS)
  Workshop}, 2011.

\bibitem{radford2021learning}
Alec Radford, Jong~Wook Kim, Chris Hallacy, Aditya Ramesh, Gabriel Goh,
  Sandhini Agarwal, Girish Sastry, Amanda Askell, Pamela Mishkin, Jack Clark,
  Gretchen Krueger, and Ilya Sutskever.
\newblock Learning transferable visual models from natural language
  supervision.
\newblock In {\em International Conference on Machine Learning (ICML)}, 2021.

\bibitem{ren2019likelihood}
Jie Ren, Peter~J Liu, Emily Fertig, Jasper Snoek, Ryan Poplin, Mark Depristo,
  Joshua Dillon, and Balaji Lakshminarayanan.
\newblock Likelihood ratios for out-of-distribution detection.
\newblock In {\em Advances in Neural Information Processing Systems (NeurIPS)},
  2019.

\bibitem{roady2020largescale}
Ryne Roady, Tyler~L. Hayes, Ronald Kemker, Ayesha Gonzales, and Christopher
  Kanan.
\newblock Are open set classification methods effective on large-scale
  datasets?
\newblock {\em PLOS ONE}, 15(9), 2020.

\bibitem{rudd2017evm}
Ethan~M. Rudd, Lalit~P. Jain, Walter~J. Scheirer, and Terrance~E. Boult.
\newblock The extreme value machine.
\newblock {\em Transactions on Pattern Analysis and Machine Intelligence
  (TPAMI)}, 2017.

\bibitem{russakovsky2015imagenet}
Olga Russakovsky, Jia Deng, Hao Su, Jonathan Krause, Sanjeev Satheesh, Sean Ma,
  Zhiheng Huang, Andrej Karpathy, Aditya Khosla, Michael Bernstein, et~al.
\newblock Imagenet large scale visual recognition challenge.
\newblock {\em International Journal of Computer Vision (IJCV)}, 115(3), 2015.

\bibitem{scheirer2013toward}
Walter~J. Scheirer, Anderson de Rezende~Rocha, Archana Sapkota, and Terrance~E.
  Boult.
\newblock Towards open set recognition.
\newblock {\em Transactions on Pattern Analysis and Machine Intelligence
  (TPAMI)}, 35(7), 2013.

\bibitem{vaze2022openset}
Sagar Vaze, Kai Han, Andrea Vedaldi, and Andrew Zissermann.
\newblock Open-set recognition: A good closed-set classifier is all you need?
\newblock In {\em International Conference on Learning Representations (ICLR)},
  2022.

\bibitem{yu2017opencategory}
Yang Yu, Wei-Yang Qu, Nan Li, and Zimin Guo.
\newblock Open-category classification by adversarial sample generation.
\newblock In {\em International Joint Conference on Artificial Intelligence
  (IJCAI)}, 2017.

\bibitem{zhou2021placeholders}
Da-Wei Zhou, Han-Jia Ye, and De-Chuan Zhan.
\newblock Learning placeholders for open-set recognition.
\newblock In {\em Conference on Computer Vision and Pattern Recognition
  (CVPR)}, 2021.

\end{thebibliography}
}

\newpage
\begin{onecolumn}
  \begin{center}
    \Huge\bf Supplemental Material\\[1ex]
  \end{center}
\end{onecolumn}

\section{Linear OSCR Curves}

Depending on the application, different false positive values are of importance.
In the main paper, we focus more on the low FPR regions since these are of more importance when a human needs to look into the positive cases and, hence, large numbers of false positives involve labor cost.
When false positives are not of utmost importance, e.g., when handled by an automated system in low-security applications, the focus will be more on having high correct classification rate.
Therefore, in \fig{oscr_all_lin} we provide the OSCR plots according to figure 2 of the main paper, but this time with linear FPR axis.
These plots highlight even more that the Entropic Open-Set (EOS) and the SoftMax with Background class (BG) approaches fail when known and unknown classes are very similar, as provided in \p2 and \p3.

\begin{figure*}[t]
	\begin{center}
	\includegraphics[width=\linewidth]{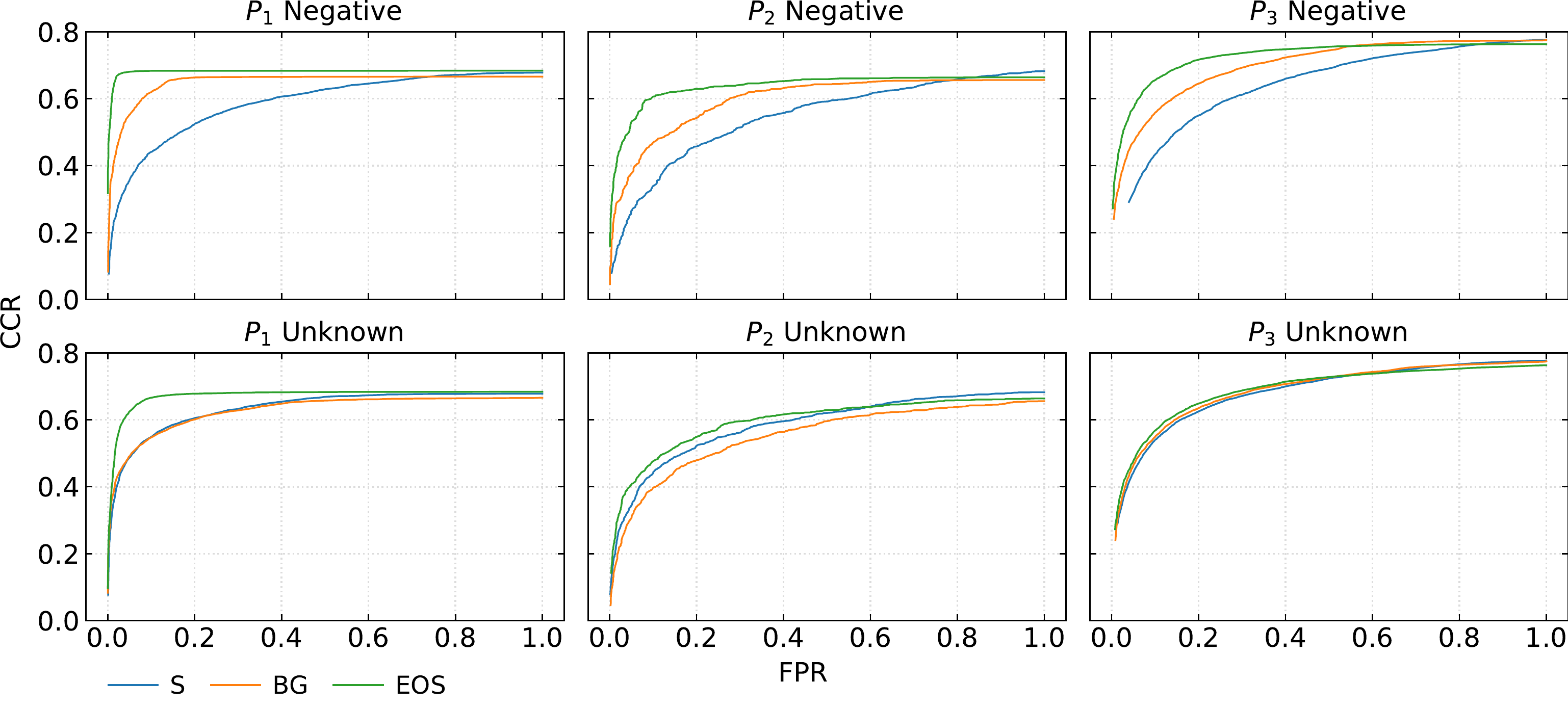}
	\end{center}
	\Caption[fig:oscr_all_lin]{Linear OSCR Curves}{
	OSCR curves in linear scale for negative data (top tow) and unknown data (bottom row) of the test set for each protocol.
	}
\end{figure*}

\section{Results Sorted by Protocol}
\fig{oscr_all_by_loss} displays the same results as shown in Figure 2 of the main paper, but here we compare the results of the different protocols.
This shows that the definition of our protocols follows our intuition.
We can see that protocol \p1 is easy in open-set, but difficult in closed-set operation, especially when looking to the performance of BG and EOS, which are designed for open-set classification.
On the opposite, protocol \p3 achieves higher closed-set recognition performance (at FPR $=1$) while dropping much faster for lower FPR values.
Protocol \p2 is difficult both in closed- and open-set evaluation, but open-set performance does not drop that drastically as in \p3, especially considering the unknown test data in the second row.
Due to its relatively small size, \p2 is optimal for hyperparameter optimization and hyperparameters should be able to be transferred to the other two protocols -- unless the algorithm is critical w.r.t. hyperparameters.
We leave hyperparameter optimization to future work.

\begin{figure*}[t]
	\begin{center}
	\includegraphics[width=\linewidth]{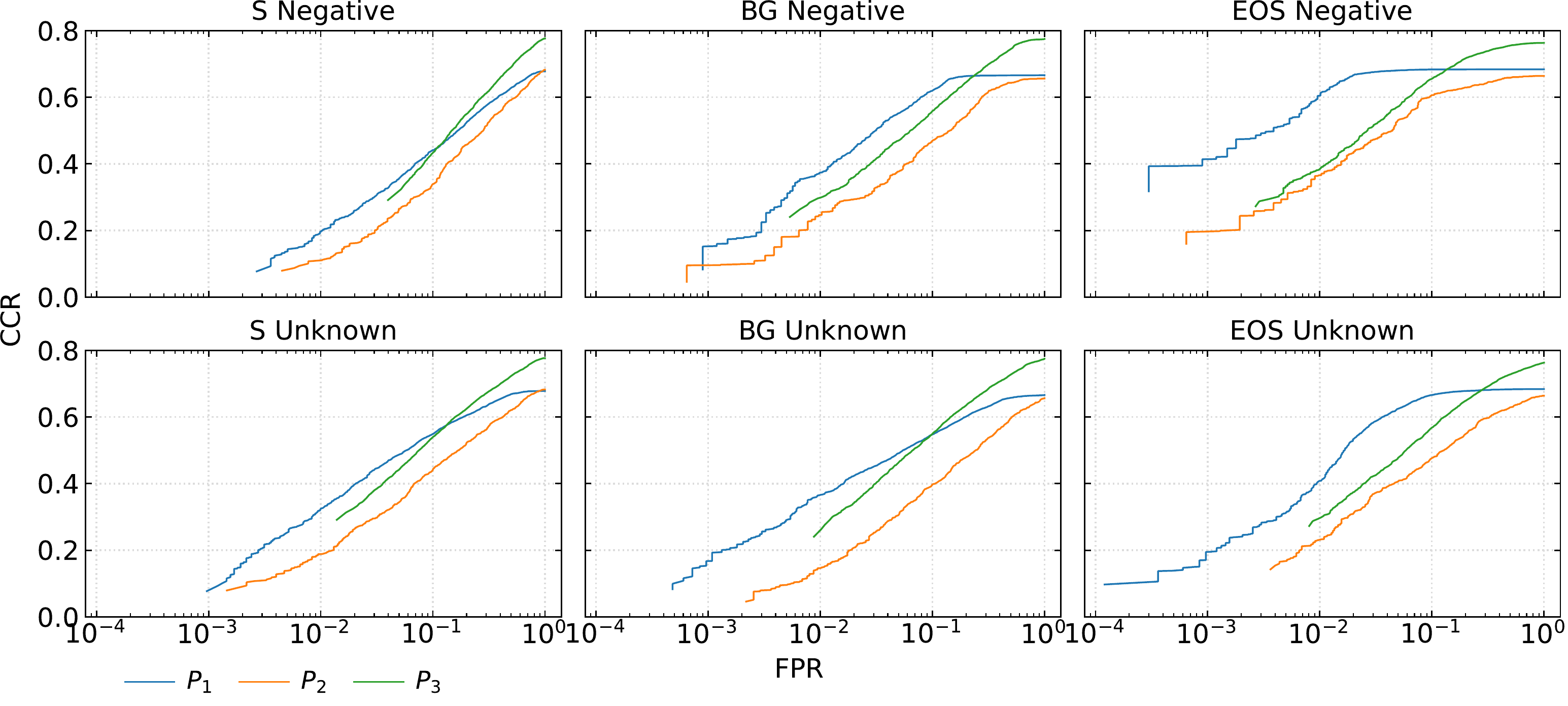}
	\end{center}
	\Caption[fig:oscr_all_by_loss]{Comparison of Protocols}{
		OSCR curves are split by algorithm to allow a comparison across protocols.
		Results are shown for negative data (top row) and unknown data (bottom row).
	}
\end{figure*}

\section{Detailed Results}
In \tab{results} we include detailed numbers of the values seen in figures 2 and 3 of the main paper.
We also include the cases where we use our confidence metric to select the best epoch of training.

When comparing \stab{results:last} with \stab{results:best} it can be seen that the selection of the best algorithm based on the confidence score provides better CCR values for BG in protocols \p1 and \p2, and better or comparable numbers for EOS across all protocols.
The advantage of S is not that expressed since S does not take into consideration negative samples and, thus, the confidence for unknown $\gamma^-$ gets very low in \stab{results:last}.
When using our validation metric, training stops early and does not provide a good overall, i.e., closed-set performance.
Therefore, we propose to use our novel evaluation metric for training open-set classifiers, but not for closed-set classifiers such as S.

\begin{table*}[t]
	\Caption[tab:results]{Classification Results}{
		Correct Classification Rates (CCR) are provided at some False Positive Rates (FPR) in the test set.
		The best CCR values are in bold.
		$\gamma^+$ is calculated using known and $\gamma^-$ using unknown classes.
	}
	\centering
	\subfloat[\label{tab:results:last}Using Last Epoch]{
		\begin{tabular}{l@{\hspace*{2em}}c@{\hspace*{1em}}cc@{\hspace*{2em}}cccc}
		\toprule
					&\multirow{2}{*}{Epoch}&	\multicolumn{2}{c@{\hspace*{2em}}}{Confidence}	& \multicolumn{4}{c}{CCR at FPR of:} \\
					\cmidrule(l{6pt}r{2em}){3-4} \cmidrule(l){5-8}
					& &$\gamma^+$& $\gamma^-$			 & 1e-3           & 1e-2           & 1e-1            & 1     \\ \midrule
				$P_1$ - S & 120 & \bf 0.665 & 0.470 & 0.077 & 0.326 & 0.549 & 0.678\\
$P_1$ - BG & 120 & 0.650 & 0.520 & 0.146 & 0.342 & 0.519 & 0.663\\
$P_1$ - EOS & 120 & 0.663 & \bf 0.846 & \bf 0.195 & \bf 0.409 & \bf 0.667 & \bf 0.684\\
\midrule
$P_2$ - S & 120 & \bf 0.665 & 0.341 & --- & 0.189 & 0.443 & \bf 0.683\\
$P_2$ - BG & 120 & 0.598 & 0.480 & --- & 0.111 & 0.339 & 0.616\\
$P_2$ - EOS & 120 & 0.592 & \bf 0.708 & --- & \bf 0.233 & \bf 0.476 & 0.664\\
\midrule
$P_3$ - S & 120 & \bf 0.767 & 0.240 & --- & --- & 0.540 & \bf 0.777\\
$P_3$ - BG & 120 & 0.753 & 0.344 & --- & 0.241 & 0.543 & 0.764\\
$P_3$ - EOS & 120 & 0.684 & \bf 0.687 & --- & \bf 0.298 & \bf 0.568 & 0.763\\

		\end{tabular}
	}

	\subfloat[\label{tab:results:best}Using Best Epoch]{
		\begin{tabular}{l@{\hspace*{2em}}c@{\hspace*{1em}}cc@{\hspace*{2em}}cccc}
			\toprule
					&\multirow{2}{*}{Epoch}&	\multicolumn{2}{c@{\hspace*{2em}}}{Confidence}	& \multicolumn{4}{c}{CCR at FPR of:} \\
					\cmidrule(l{6pt}r{2em}){3-4} \cmidrule(l){5-8}
					& &$\gamma^+$& $\gamma^-$			 & 1e-3           & 1e-2           & 1e-1            & 1     \\ \midrule
				$P_1$ - S & 20 & 0.584 & \bf 0.906 & 0.270 & 0.543 & 0.664 & 0.667\\
$P_1$ - BG & 99 & 0.660 & 0.628 & 0.167 & 0.420 & 0.572 & 0.673\\
$P_1$ - EOS & 105 & \bf 0.672 & 0.877 & \bf 0.283 & \bf 0.535 & \bf 0.683 & \bf 0.694\\
\midrule
$P_2$ - S & 39 & 0.603 & 0.528 & \bf 0.031 & 0.134 & 0.384 & 0.659\\
$P_2$ - BG & 113 & \bf 0.638 & 0.491 & --- & 0.156 & 0.368 & 0.658\\
$P_2$ - EOS & 101 & 0.599 & \bf 0.694 & --- & \bf 0.223 & \bf 0.473 & \bf 0.666\\
\midrule
$P_3$ - S & 15 & 0.677 & 0.527 & \bf 0.083 & 0.239 & 0.509 & 0.745\\
$P_3$ - BG & 115 & \bf 0.747 & 0.379 & --- & 0.243 & 0.539 & 0.761\\
$P_3$ - EOS & 114 & 0.696 & \bf 0.688 & --- & \bf 0.266 & \bf 0.585 & \bf 0.772\\

		\end{tabular}
	}
\end{table*}

\section{Detailed List of Classes}
For reproducibility, we also provide the list of classes (ImageNet identifier and class name) that we used in our three protocols in Tables~\ref{tab:p1_classes}, \ref{tab:p2_classes} and \ref{tab:p3_classes}.
For known and negative classes, we used the samples of the original training partition for training and validation, and the samples from the validation partition for testing -- since the test set labels of ILSVRC2012 are still not publicly available.

\onecolumn
\scriptsize
\begin{longtable}{@{}llllll@{}}

\caption{\textsc{Protocol 1 Classes}: List of super-classes and leaf nodes of Protocol \p1.}
\label{tab:p1_classes}\\
\toprule
    \multicolumn{2}{c}{Known} & \multicolumn{2}{c}{Negative} & \multicolumn{2}{c}{Unknown} \\
    \cmidrule(l{2pt}r{2pt}){1-2} \cmidrule(l{2pt}r{2pt}){3-4} \cmidrule(l{2pt}r{2pt}){5-6}
       Id &                           Name &        Id &                Name &        Id &                Name \\
\midrule
\endfirsthead

\caption*{\textsc{Protocol 1 Classes}: List of super-classes and leaf nodes of Protocol \p1.}\\
\toprule
    \multicolumn{2}{c}{Known} & \multicolumn{2}{c}{Negative} & \multicolumn{2}{c}{Unknown} \\
    \cmidrule(l{2pt}r{2pt}){1-2} \cmidrule(l{2pt}r{2pt}){3-4} \cmidrule(l{2pt}r{2pt}){5-6}
       Id &                           Name &        Id &                Name &        Id &                Name \\
\midrule
\endhead

\midrule
\multicolumn{6}{c}{{Continued on next page}} \\
\midrule
\endfoot

\midrule
\multicolumn{6}{c}{{End of table Protocol 1}} \\
\bottomrule
\endlastfoot

\it n02084071 & \it dog & \it n02118333 & \it fox & \it n07555863 & \it food\\
\qquad n02085620 & \qquad Chihuahua & \qquad n02119022 & \qquad red fox & \qquad n07684084 & \qquad French loaf\\
\qquad n02085782 & \qquad Japanese spaniel & \qquad n02119789 & \qquad kit fox & \qquad n07693725 & \qquad bagel\\
\qquad n02085936 & \qquad Maltese dog & \qquad n02120079 & \qquad Arctic fox & \qquad n07695742 & \qquad pretzel\\
\qquad n02086079 & \qquad Pekinese & \qquad n02120505 & \qquad grey fox & \qquad n07714571 & \qquad head cabbage\\
\qquad n02086240 & \qquad Shih-Tzu & \it n02115335 & \it wild dog & \qquad n07714990 & \qquad broccoli\\
\qquad n02086646 & \qquad Blenheim spaniel & \qquad n02115641 & \qquad dingo & \qquad n07715103 & \qquad cauliflower\\
\qquad n02086910 & \qquad papillon & \qquad n02115913 & \qquad dhole & \qquad n07716358 & \qquad zucchini\\
\qquad n02087046 & \qquad toy terrier & \qquad n02116738 & \qquad African hunting dog & \qquad n07716906 & \qquad spaghetti squash\\
\qquad n02087394 & \qquad Rhodesian ridgeback & \it n02114100 & \it wolf & \qquad n07717410 & \qquad acorn squash\\
\qquad n02088094 & \qquad Afghan hound & \qquad n02114367 & \qquad timber wolf & \qquad n07717556 & \qquad butternut squash\\
\qquad n02088238 & \qquad basset & \qquad n02114548 & \qquad white wolf & \qquad n07718472 & \qquad cucumber\\
\qquad n02088364 & \qquad beagle & \qquad n02114712 & \qquad red wolf & \qquad n07718747 & \qquad artichoke\\
\qquad n02088466 & \qquad bloodhound & \qquad n02114855 & \qquad coyote & \qquad n07720875 & \qquad bell pepper\\
\qquad n02088632 & \qquad bluetick & \it n02120997 & \it feline & \qquad n07730033 & \qquad cardoon\\
\qquad n02089078 & \qquad black-and-tan coonho & \qquad n02125311 & \qquad cougar & \qquad n07734744 & \qquad mushroom\\
\qquad n02089867 & \qquad Walker hound & \qquad n02127052 & \qquad lynx & \qquad n07745940 & \qquad strawberry\\
\qquad n02089973 & \qquad English foxhound & \qquad n02128385 & \qquad leopard & \qquad n07747607 & \qquad orange\\
\qquad n02090379 & \qquad redbone & \qquad n02128757 & \qquad snow leopard & \qquad n07749582 & \qquad lemon\\
\qquad n02090622 & \qquad borzoi & \qquad n02128925 & \qquad jaguar & \qquad n07753113 & \qquad fig\\
\qquad n02090721 & \qquad Irish wolfhound & \qquad n02129165 & \qquad lion & \qquad n07753275 & \qquad pineapple\\
\qquad n02091244 & \qquad Ibizan hound & \qquad n02129604 & \qquad tiger & \qquad n07753592 & \qquad banana\\
\qquad n02091467 & \qquad Norwegian elkhound & \qquad n02130308 & \qquad cheetah & \qquad n07754684 & \qquad jackfruit\\
\qquad n02091635 & \qquad otterhound & \it n02131653 & \it bear & \qquad n07760859 & \qquad custard apple\\
\qquad n02091831 & \qquad Saluki & \qquad n02132136 & \qquad brown bear & \qquad n07768694 & \qquad pomegranate\\
\qquad n02092002 & \qquad Scottish deerhound & \qquad n02133161 & \qquad American black bear & \it n03791235 & \it motor vehicle\\
\qquad n02092339 & \qquad Weimaraner & \qquad n02134084 & \qquad ice bear & \qquad n02701002 & \qquad ambulance\\
\qquad n02093256 & \qquad Staffordshire bullte & \qquad n02134418 & \qquad sloth bear & \qquad n02704792 & \qquad amphibian\\
\qquad n02093428 & \qquad American Staffordshi & \it n02441326 & \it musteline mammal & \qquad n02814533 & \qquad beach wagon\\
\qquad n02093647 & \qquad Bedlington terrier & \qquad n02441942 & \qquad weasel & \qquad n02930766 & \qquad cab\\
\qquad n02093754 & \qquad Border terrier & \qquad n02442845 & \qquad mink & \qquad n03100240 & \qquad convertible\\
\qquad n02093859 & \qquad Kerry blue terrier & \qquad n02443114 & \qquad polecat & \qquad n03345487 & \qquad fire engine\\
\qquad n02093991 & \qquad Irish terrier & \qquad n02443484 & \qquad black-footed ferret & \qquad n03417042 & \qquad garbage truck\\
\qquad n02094114 & \qquad Norfolk terrier & \qquad n02444819 & \qquad otter & \qquad n03444034 & \qquad go-kart\\
\qquad n02094258 & \qquad Norwich terrier & \qquad n02445715 & \qquad skunk & \qquad n03594945 & \qquad jeep\\
\qquad n02094433 & \qquad Yorkshire terrier & \qquad n02447366 & \qquad badger & \qquad n03670208 & \qquad limousine\\
\qquad n02095314 & \qquad wire-haired fox terr & \it n02370806 & \it ungulate & \qquad n03770679 & \qquad minivan\\
\qquad n02095570 & \qquad Lakeland terrier & \qquad n02389026 & \qquad sorrel & \qquad n03777568 & \qquad Model T\\
\qquad n02095889 & \qquad Sealyham terrier & \qquad n02391049 & \qquad zebra & \qquad n03785016 & \qquad moped\\
\qquad n02096051 & \qquad Airedale & \qquad n02395406 & \qquad hog & \qquad n03796401 & \qquad moving van\\
\qquad n02096177 & \qquad cairn & \qquad n02396427 & \qquad wild boar & \qquad n03930630 & \qquad pickup\\
\qquad n02096294 & \qquad Australian terrier & \qquad n02397096 & \qquad warthog & \qquad n03977966 & \qquad police van\\
\qquad n02096437 & \qquad Dandie Dinmont & \qquad n02398521 & \qquad hippopotamus & \qquad n04037443 & \qquad racer\\
\qquad n02096585 & \qquad Boston bull & \qquad n02403003 & \qquad ox & \qquad n04252225 & \qquad snowplow\\
\qquad n02097047 & \qquad miniature schnauzer & \qquad n02408429 & \qquad water buffalo & \qquad n04285008 & \qquad sports car\\
\qquad n02097130 & \qquad giant schnauzer & \qquad n02410509 & \qquad bison & \qquad n04461696 & \qquad tow truck\\
\qquad n02097209 & \qquad standard schnauzer & \qquad n02412080 & \qquad ram & \qquad n04467665 & \qquad trailer truck\\
\qquad n02097298 & \qquad Scotch terrier & \qquad n02415577 & \qquad bighorn & \it n03183080 & \it device\\
\qquad n02097474 & \qquad Tibetan terrier & \qquad n02417914 & \qquad ibex & \qquad n02666196 & \qquad abacus\\
\qquad n02097658 & \qquad silky terrier & \qquad n02422106 & \qquad hartebeest & \qquad n02672831 & \qquad accordion\\
\qquad n02098105 & \qquad soft-coated wheaten  & \qquad n02422699 & \qquad impala & \qquad n02676566 & \qquad acoustic guitar\\
\qquad n02098286 & \qquad West Highland white  & \qquad n02423022 & \qquad gazelle & \qquad n02708093 & \qquad analog clock\\
\qquad n02098413 & \qquad Lhasa & \qquad n02437312 & \qquad Arabian camel & \qquad n02749479 & \qquad assault rifle\\
\qquad n02099267 & \qquad flat-coated retrieve & \qquad n02437616 & \qquad llama & \qquad n02787622 & \qquad banjo\\
\qquad n02099429 & \qquad curly-coated retriev & \it n02469914 & \it primate & \qquad n02794156 & \qquad barometer\\
\qquad n02099601 & \qquad golden retriever & \qquad n02480495 & \qquad orangutan & \qquad n02804610 & \qquad bassoon\\
\qquad n02099712 & \qquad Labrador retriever & \qquad n02480855 & \qquad gorilla & \qquad n02841315 & \qquad binoculars\\
\qquad n02099849 & \qquad Chesapeake Bay retri & \qquad n02481823 & \qquad chimpanzee & \qquad n02879718 & \qquad bow\\
\qquad n02100236 & \qquad German short-haired  & \qquad n02483362 & \qquad gibbon & \qquad n02910353 & \qquad buckle\\
\qquad n02100583 & \qquad vizsla & \qquad n02483708 & \qquad siamang & \qquad n02948072 & \qquad candle\\
\qquad n02100735 & \qquad English setter & \qquad n02484975 & \qquad guenon & \qquad n02950826 & \qquad cannon\\
\qquad n02100877 & \qquad Irish setter & \qquad n02486261 & \qquad patas & \qquad n02965783 & \qquad car mirror\\
\qquad n02101006 & \qquad Gordon setter & \qquad n02486410 & \qquad baboon & \qquad n02966193 & \qquad carousel\\
\qquad n02101388 & \qquad Brittany spaniel & \qquad n02487347 & \qquad macaque & \qquad n02974003 & \qquad car wheel\\
\qquad n02101556 & \qquad clumber & \qquad n02488291 & \qquad langur & \qquad n02977058 & \qquad cash machine\\
\qquad n02102040 & \qquad English springer & \qquad n02488702 & \qquad colobus & \qquad n02992211 & \qquad cello\\
\qquad n02102177 & \qquad Welsh springer spani & \qquad n02489166 & \qquad proboscis monkey & \qquad n03000684 & \qquad chain saw\\
\qquad n02102318 & \qquad cocker spaniel & \qquad n02490219 & \qquad marmoset & \qquad n03017168 & \qquad chime\\
\qquad n02102480 & \qquad Sussex spaniel & \qquad n02492035 & \qquad capuchin & \qquad n03075370 & \qquad combination lock\\
\qquad n02102973 & \qquad Irish water spaniel & \qquad n02492660 & \qquad howler monkey & \qquad n03110669 & \qquad cornet\\
\qquad n02104029 & \qquad kuvasz & \qquad n02493509 & \qquad titi & \qquad n03126707 & \qquad crane\\
\qquad n02104365 & \qquad schipperke & \qquad n02493793 & \qquad spider monkey & \qquad n03180011 & \qquad desktop computer\\
\qquad n02105056 & \qquad groenendael & \qquad n02494079 & \qquad squirrel monkey & \qquad n03196217 & \qquad digital clock\\
\qquad n02105162 & \qquad malinois & \qquad n02497673 & \qquad Madagascar cat & \qquad n03197337 & \qquad digital watch\\
\qquad n02105251 & \qquad briard & \qquad n02500267 & \qquad indri & \qquad n03208938 & \qquad disk brake\\
\qquad n02105412 & \qquad kelpie &  &  & \qquad n03249569 & \qquad drum\\
\qquad n02105505 & \qquad komondor &  &  & \qquad n03271574 & \qquad electric fan\\
\qquad n02105641 & \qquad Old English sheepdog &  &  & \qquad n03272010 & \qquad electric guitar\\
\qquad n02105855 & \qquad Shetland sheepdog &  &  & \qquad n03372029 & \qquad flute\\
\qquad n02106030 & \qquad collie &  &  & \qquad n03394916 & \qquad French horn\\
\qquad n02106166 & \qquad Border collie &  &  & \qquad n03425413 & \qquad gas pump\\
\qquad n02106382 & \qquad Bouvier des Flandres &  &  & \qquad n03447721 & \qquad gong\\
\qquad n02106550 & \qquad Rottweiler &  &  & \qquad n03452741 & \qquad grand piano\\
\qquad n02106662 & \qquad German shepherd &  &  & \qquad n03467068 & \qquad guillotine\\
\qquad n02107142 & \qquad Doberman &  &  & \qquad n03476684 & \qquad hair slide\\
\qquad n02107312 & \qquad miniature pinscher &  &  & \qquad n03485407 & \qquad hand-held computer\\
\qquad n02107574 & \qquad Greater Swiss Mounta &  &  & \qquad n03492542 & \qquad hard disc\\
\qquad n02107683 & \qquad Bernese mountain dog &  &  & \qquad n03494278 & \qquad harmonica\\
\qquad n02107908 & \qquad Appenzeller &  &  & \qquad n03495258 & \qquad harp\\
\qquad n02108000 & \qquad EntleBucher &  &  & \qquad n03496892 & \qquad harvester\\
\qquad n02108089 & \qquad boxer &  &  & \qquad n03532672 & \qquad hook\\
\qquad n02108422 & \qquad bull mastiff &  &  & \qquad n03544143 & \qquad hourglass\\
\qquad n02108551 & \qquad Tibetan mastiff &  &  & \qquad n03590841 & \qquad jack-o'-lantern\\
\qquad n02108915 & \qquad French bulldog &  &  & \qquad n03627232 & \qquad knot\\
\qquad n02109047 & \qquad Great Dane &  &  & \qquad n03642806 & \qquad laptop\\
\qquad n02109525 & \qquad Saint Bernard &  &  & \qquad n03666591 & \qquad lighter\\
\qquad n02109961 & \qquad Eskimo dog &  &  & \qquad n03691459 & \qquad loudspeaker\\
\qquad n02110063 & \qquad malamute &  &  & \qquad n03692522 & \qquad loupe\\
\qquad n02110185 & \qquad Siberian husky &  &  & \qquad n03706229 & \qquad magnetic compass\\
\qquad n02110341 & \qquad dalmatian &  &  & \qquad n03720891 & \qquad maraca\\
\qquad n02110627 & \qquad affenpinscher &  &  & \qquad n03721384 & \qquad marimba\\
\qquad n02110806 & \qquad basenji &  &  & \qquad n03733131 & \qquad maypole\\
\qquad n02110958 & \qquad pug &  &  & \qquad n03759954 & \qquad microphone\\
\qquad n02111129 & \qquad Leonberg &  &  & \qquad n03773504 & \qquad missile\\
\qquad n02111277 & \qquad Newfoundland &  &  & \qquad n03793489 & \qquad mouse\\
\qquad n02111500 & \qquad Great Pyrenees &  &  & \qquad n03794056 & \qquad mousetrap\\
\qquad n02111889 & \qquad Samoyed &  &  & \qquad n03803284 & \qquad muzzle\\
\qquad n02112018 & \qquad Pomeranian &  &  & \qquad n03804744 & \qquad nail\\
\qquad n02112137 & \qquad chow &  &  & \qquad n03814639 & \qquad neck brace\\
\qquad n02112350 & \qquad keeshond &  &  & \qquad n03832673 & \qquad notebook\\
\qquad n02112706 & \qquad Brabancon griffon &  &  & \qquad n03838899 & \qquad oboe\\
\qquad n02113023 & \qquad Pembroke &  &  & \qquad n03840681 & \qquad ocarina\\
\qquad n02113186 & \qquad Cardigan &  &  & \qquad n03841143 & \qquad odometer\\
\qquad n02113624 & \qquad toy poodle &  &  & \qquad n03843555 & \qquad oil filter\\
\qquad n02113712 & \qquad miniature poodle &  &  & \qquad n03854065 & \qquad organ\\
\qquad n02113799 & \qquad standard poodle &  &  & \qquad n03868863 & \qquad oxygen mask\\
\qquad n02113978 & \qquad Mexican hairless &  &  & \qquad n03874293 & \qquad paddlewheel\\
 &  &  &  & \qquad n03874599 & \qquad padlock\\
 &  &  &  & \qquad n03884397 & \qquad panpipe\\
 &  &  &  & \qquad n03891332 & \qquad parking meter\\
 &  &  &  & \qquad n03929660 & \qquad pick\\
 &  &  &  & \qquad n03933933 & \qquad pier\\
 &  &  &  & \qquad n03944341 & \qquad pinwheel\\
 &  &  &  & \qquad n03992509 & \qquad potter's wheel\\
 &  &  &  & \qquad n03995372 & \qquad power drill\\
 &  &  &  & \qquad n04008634 & \qquad projectile\\
 &  &  &  & \qquad n04009552 & \qquad projector\\
 &  &  &  & \qquad n04040759 & \qquad radiator\\
 &  &  &  & \qquad n04044716 & \qquad radio telescope\\
 &  &  &  & \qquad n04067472 & \qquad reel\\
 &  &  &  & \qquad n04074963 & \qquad remote control\\
 &  &  &  & \qquad n04086273 & \qquad revolver\\
 &  &  &  & \qquad n04090263 & \qquad rifle\\
 &  &  &  & \qquad n04118776 & \qquad rule\\
 &  &  &  & \qquad n04127249 & \qquad safety pin\\
 &  &  &  & \qquad n04141076 & \qquad sax\\
 &  &  &  & \qquad n04141975 & \qquad scale\\
 &  &  &  & \qquad n04152593 & \qquad screen\\
 &  &  &  & \qquad n04153751 & \qquad screw\\
 &  &  &  & \qquad n04228054 & \qquad ski\\
 &  &  &  & \qquad n04238763 & \qquad slide rule\\
 &  &  &  & \qquad n04243546 & \qquad slot\\
 &  &  &  & \qquad n04251144 & \qquad snorkel\\
 &  &  &  & \qquad n04258138 & \qquad solar dish\\
 &  &  &  & \qquad n04265275 & \qquad space heater\\
 &  &  &  & \qquad n04275548 & \qquad spider web\\
 &  &  &  & \qquad n04286575 & \qquad spotlight\\
 &  &  &  & \qquad n04311174 & \qquad steel drum\\
 &  &  &  & \qquad n04317175 & \qquad stethoscope\\
 &  &  &  & \qquad n04328186 & \qquad stopwatch\\
 &  &  &  & \qquad n04330267 & \qquad stove\\
 &  &  &  & \qquad n04332243 & \qquad strainer\\
 &  &  &  & \qquad n04355338 & \qquad sundial\\
 &  &  &  & \qquad n04355933 & \qquad sunglass\\
 &  &  &  & \qquad n04356056 & \qquad sunglasses\\
 &  &  &  & \qquad n04372370 & \qquad switch\\
 &  &  &  & \qquad n04376876 & \qquad syringe\\
 &  &  &  & \qquad n04428191 & \qquad thresher\\
 &  &  &  & \qquad n04456115 & \qquad torch\\
 &  &  &  & \qquad n04485082 & \qquad tripod\\
 &  &  &  & \qquad n04487394 & \qquad trombone\\
 &  &  &  & \qquad n04505470 & \qquad typewriter keyboard\\
 &  &  &  & \qquad n04515003 & \qquad upright\\
 &  &  &  & \qquad n04525305 & \qquad vending machine\\
 &  &  &  & \qquad n04536866 & \qquad violin\\
 &  &  &  & \qquad n04548280 & \qquad wall clock\\
 &  &  &  & \qquad n04579432 & \qquad whistle\\
 &  &  &  & \qquad n04592741 & \qquad wing\\
 &  &  &  & \qquad n06359193 & \qquad web site\\

\end{longtable}

\newpage
\begin{longtable}{@{}llllll@{}}

\caption{\textsc{Protocol 2 Classes}: List of super-classes and leaf nodes of Protocol \p2.}
\label{tab:p2_classes}\\
\toprule
    \multicolumn{2}{c}{Known} & \multicolumn{2}{c}{Negative} & \multicolumn{2}{c}{Unknown} \\
    \cmidrule(l{2pt}r{2pt}){1-2} \cmidrule(l{2pt}r{2pt}){3-4} \cmidrule(l{2pt}r{2pt}){5-6}
       Id &                           Name &        Id &                Name &        Id &                Name \\
\midrule
\endfirsthead

\caption*{\textsc{Protocol 2 Classes}: List of super-classes and leaf nodes of Protocol \p2.}\\
\toprule
    \multicolumn{2}{c}{Known} & \multicolumn{2}{c}{Negative} & \multicolumn{2}{c}{Unknown} \\
    \cmidrule(l{2pt}r{2pt}){1-2} \cmidrule(l{2pt}r{2pt}){3-4} \cmidrule(l{2pt}r{2pt}){5-6}
       Id &                           Name &        Id &                Name &        Id &                Name \\
\midrule
\endhead

\midrule
\multicolumn{6}{c}{{Continued on next page}} \\
\midrule
\endfoot

\midrule
\multicolumn{6}{c}{{End of table Protocol 2}} \\
\bottomrule
\endlastfoot

\it n02087122 & \it hunting dog & \it n02087122 & \it hunting dog & \it n02085374 & \it toy dog\\
\qquad n02087394 & \qquad Rhodesian ridgeback & \qquad n02096051 & \qquad Airedale & \qquad n02085620 & \qquad Chihuahua\\
\qquad n02088094 & \qquad Afghan hound & \qquad n02096177 & \qquad cairn & \qquad n02085782 & \qquad Japanese spaniel\\
\qquad n02088238 & \qquad basset & \qquad n02096294 & \qquad Australian terrier & \qquad n02085936 & \qquad Maltese dog\\
\qquad n02088364 & \qquad beagle & \qquad n02096437 & \qquad Dandie Dinmont & \qquad n02086079 & \qquad Pekinese\\
\qquad n02088466 & \qquad bloodhound & \qquad n02096585 & \qquad Boston bull & \qquad n02086240 & \qquad Shih-Tzu\\
\qquad n02088632 & \qquad bluetick & \qquad n02097047 & \qquad miniature schnauzer & \qquad n02086646 & \qquad Blenheim spaniel\\
\qquad n02089078 & \qquad black-and-tan coonho & \qquad n02097130 & \qquad giant schnauzer & \qquad n02086910 & \qquad papillon\\
\qquad n02089867 & \qquad Walker hound & \qquad n02097209 & \qquad standard schnauzer & \qquad n02087046 & \qquad toy terrier\\
\qquad n02089973 & \qquad English foxhound & \qquad n02097298 & \qquad Scotch terrier & \it n02118333 & \it fox\\
\qquad n02090379 & \qquad redbone & \qquad n02097474 & \qquad Tibetan terrier & \qquad n02119022 & \qquad red fox\\
\qquad n02090622 & \qquad borzoi & \qquad n02097658 & \qquad silky terrier & \qquad n02119789 & \qquad kit fox\\
\qquad n02090721 & \qquad Irish wolfhound & \qquad n02098105 & \qquad soft-coated wheaten  & \qquad n02120079 & \qquad Arctic fox\\
\qquad n02091244 & \qquad Ibizan hound & \qquad n02098286 & \qquad West Highland white  & \qquad n02120505 & \qquad grey fox\\
\qquad n02091467 & \qquad Norwegian elkhound & \qquad n02098413 & \qquad Lhasa & \it n02115335 & \it wild dog\\
\qquad n02091635 & \qquad otterhound & \qquad n02099267 & \qquad flat-coated retrieve & \qquad n02115641 & \qquad dingo\\
\qquad n02091831 & \qquad Saluki & \qquad n02099429 & \qquad curly-coated retriev & \qquad n02115913 & \qquad dhole\\
\qquad n02092002 & \qquad Scottish deerhound & \qquad n02099601 & \qquad golden retriever & \qquad n02116738 & \qquad African hunting dog\\
\qquad n02092339 & \qquad Weimaraner & \qquad n02099712 & \qquad Labrador retriever & \it n02114100 & \it wolf\\
\qquad n02093256 & \qquad Staffordshire bullte & \qquad n02099849 & \qquad Chesapeake Bay retri & \qquad n02114367 & \qquad timber wolf\\
\qquad n02093428 & \qquad American Staffordshi & \qquad n02100236 & \qquad German short-haired  & \qquad n02114548 & \qquad white wolf\\
\qquad n02093647 & \qquad Bedlington terrier & \qquad n02100583 & \qquad vizsla & \qquad n02114712 & \qquad red wolf\\
\qquad n02093754 & \qquad Border terrier & \qquad n02100735 & \qquad English setter & \qquad n02114855 & \qquad coyote\\
\qquad n02093859 & \qquad Kerry blue terrier & \qquad n02100877 & \qquad Irish setter & \it n02120997 & \it feline\\
\qquad n02093991 & \qquad Irish terrier & \qquad n02101006 & \qquad Gordon setter & \qquad n02125311 & \qquad cougar\\
\qquad n02094114 & \qquad Norfolk terrier & \qquad n02101388 & \qquad Brittany spaniel & \qquad n02127052 & \qquad lynx\\
\qquad n02094258 & \qquad Norwich terrier & \qquad n02101556 & \qquad clumber & \qquad n02128385 & \qquad leopard\\
\qquad n02094433 & \qquad Yorkshire terrier & \qquad n02102040 & \qquad English springer & \qquad n02128757 & \qquad snow leopard\\
\qquad n02095314 & \qquad wire-haired fox terr & \qquad n02102177 & \qquad Welsh springer spani & \qquad n02128925 & \qquad jaguar\\
\qquad n02095570 & \qquad Lakeland terrier & \qquad n02102318 & \qquad cocker spaniel & \qquad n02129165 & \qquad lion\\
\qquad n02095889 & \qquad Sealyham terrier & \qquad n02102480 & \qquad Sussex spaniel & \qquad n02129604 & \qquad tiger\\
 &  & \qquad n02102973 & \qquad Irish water spaniel & \qquad n02130308 & \qquad cheetah\\
 &  &  &  & \it n02131653 & \it bear\\
 &  &  &  & \qquad n02132136 & \qquad brown bear\\
 &  &  &  & \qquad n02133161 & \qquad American black bear\\
 &  &  &  & \qquad n02134084 & \qquad ice bear\\
 &  &  &  & \qquad n02134418 & \qquad sloth bear\\
 &  &  &  & \it n02441326 & \it musteline mammal\\
 &  &  &  & \qquad n02441942 & \qquad weasel\\
 &  &  &  & \qquad n02442845 & \qquad mink\\
 &  &  &  & \qquad n02443114 & \qquad polecat\\
 &  &  &  & \qquad n02443484 & \qquad black-footed ferret\\
 &  &  &  & \qquad n02444819 & \qquad otter\\
 &  &  &  & \qquad n02445715 & \qquad skunk\\
 &  &  &  & \qquad n02447366 & \qquad badger\\
 &  &  &  & \it n02370806 & \it ungulate\\
 &  &  &  & \qquad n02389026 & \qquad sorrel\\
 &  &  &  & \qquad n02391049 & \qquad zebra\\
 &  &  &  & \qquad n02395406 & \qquad hog\\
 &  &  &  & \qquad n02396427 & \qquad wild boar\\
 &  &  &  & \qquad n02397096 & \qquad warthog\\
 &  &  &  & \qquad n02398521 & \qquad hippopotamus\\
 &  &  &  & \qquad n02403003 & \qquad ox\\
 &  &  &  & \qquad n02408429 & \qquad water buffalo\\
 &  &  &  & \qquad n02410509 & \qquad bison\\
 &  &  &  & \qquad n02412080 & \qquad ram\\
 &  &  &  & \qquad n02415577 & \qquad bighorn\\
 &  &  &  & \qquad n02417914 & \qquad ibex\\
 &  &  &  & \qquad n02422106 & \qquad hartebeest\\
 &  &  &  & \qquad n02422699 & \qquad impala\\
 &  &  &  & \qquad n02423022 & \qquad gazelle\\
 &  &  &  & \qquad n02437312 & \qquad Arabian camel\\
 &  &  &  & \qquad n02437616 & \qquad llama\\

\end{longtable}

\newpage
\begin{longtable}{@{}llllll@{}}

\caption{\textsc{Protocol 3 Classes}: List of super-classes and leaf nodes of Protocol \p3.}
\label{tab:p3_classes}\\
\toprule
    \multicolumn{2}{c}{Known} & \multicolumn{2}{c}{Negative} & \multicolumn{2}{c}{Unknown} \\
    \cmidrule(l{2pt}r{2pt}){1-2} \cmidrule(l{2pt}r{2pt}){3-4} \cmidrule(l{2pt}r{2pt}){5-6}
       Id &                           Name &        Id &                Name &        Id &                Name \\
\midrule
\endfirsthead

\caption*{\textsc{Protocol 3 Classes}: List of super-classes and leaf nodes of Protocol \p3.}\\
\toprule
    \multicolumn{2}{c}{Known} & \multicolumn{2}{c}{Negative} & \multicolumn{2}{c}{Unknown} \\
    \cmidrule(l{2pt}r{2pt}){1-2} \cmidrule(l{2pt}r{2pt}){3-4} \cmidrule(l{2pt}r{2pt}){5-6}
       Id &                           Name &        Id &                Name &        Id &                Name \\
\midrule
\endhead

\midrule
\multicolumn{6}{c}{{Continued on next page}} \\
\midrule
\endfoot

\midrule
\multicolumn{6}{c}{{End of table Protocol 3}} \\
\bottomrule
\endlastfoot

\it n02084071 & \it dog & \it n02084071 & \it dog & \it n02084071 & \it dog\\
\qquad n02085620 & \qquad Chihuahua & \qquad n02085782 & \qquad Japanese spaniel & \qquad n02086079 & \qquad Pekinese\\
\qquad n02085936 & \qquad Maltese dog & \qquad n02086646 & \qquad Blenheim spaniel & \qquad n02088094 & \qquad Afghan hound\\
\qquad n02086240 & \qquad Shih-Tzu & \qquad n02087046 & \qquad toy terrier & \qquad n02089867 & \qquad Walker hound\\
\qquad n02086910 & \qquad papillon & \qquad n02088364 & \qquad beagle & \qquad n02091467 & \qquad Norwegian elkhound\\
\qquad n02087394 & \qquad Rhodesian ridgeback & \qquad n02088632 & \qquad bluetick & \qquad n02093428 & \qquad American Staffordshi\\
\qquad n02088238 & \qquad basset & \qquad n02090379 & \qquad redbone & \qquad n02094258 & \qquad Norwich terrier\\
\qquad n02088466 & \qquad bloodhound & \qquad n02090721 & \qquad Irish wolfhound & \qquad n02096177 & \qquad cairn\\
\qquad n02089078 & \qquad black-and-tan coonho & \qquad n02091831 & \qquad Saluki & \qquad n02097209 & \qquad standard schnauzer\\
\qquad n02089973 & \qquad English foxhound & \qquad n02092339 & \qquad Weimaraner & \qquad n02098413 & \qquad Lhasa\\
\qquad n02090622 & \qquad borzoi & \qquad n02093754 & \qquad Border terrier & \qquad n02100236 & \qquad German short-haired \\
\qquad n02091244 & \qquad Ibizan hound & \qquad n02093991 & \qquad Irish terrier & \qquad n02101556 & \qquad clumber\\
\qquad n02091635 & \qquad otterhound & \qquad n02095314 & \qquad wire-haired fox terr & \qquad n02104029 & \qquad kuvasz\\
\qquad n02092002 & \qquad Scottish deerhound & \qquad n02095889 & \qquad Sealyham terrier & \qquad n02105505 & \qquad komondor\\
\qquad n02093256 & \qquad Staffordshire bullte & \qquad n02096437 & \qquad Dandie Dinmont & \qquad n02106550 & \qquad Rottweiler\\
\qquad n02093647 & \qquad Bedlington terrier & \qquad n02097047 & \qquad miniature schnauzer & \qquad n02107908 & \qquad Appenzeller\\
\qquad n02093859 & \qquad Kerry blue terrier & \qquad n02097474 & \qquad Tibetan terrier & \qquad n02109047 & \qquad Great Dane\\
\qquad n02094114 & \qquad Norfolk terrier & \qquad n02098105 & \qquad soft-coated wheaten  & \qquad n02110627 & \qquad affenpinscher\\
\qquad n02094433 & \qquad Yorkshire terrier & \qquad n02099429 & \qquad curly-coated retriev & \qquad n02111889 & \qquad Samoyed\\
\qquad n02095570 & \qquad Lakeland terrier & \qquad n02099712 & \qquad Labrador retriever & \qquad n02113186 & \qquad Cardigan\\
\qquad n02096051 & \qquad Airedale & \qquad n02100735 & \qquad English setter & \it n01503061 & \it bird\\
\qquad n02096294 & \qquad Australian terrier & \qquad n02101006 & \qquad Gordon setter & \qquad n01530575 & \qquad brambling\\
\qquad n02096585 & \qquad Boston bull & \qquad n02102177 & \qquad Welsh springer spani & \qquad n01560419 & \qquad bulbul\\
\qquad n02097130 & \qquad giant schnauzer & \qquad n02102480 & \qquad Sussex spaniel & \qquad n01614925 & \qquad bald eagle\\
\qquad n02097298 & \qquad Scotch terrier & \qquad n02105056 & \qquad groenendael & \qquad n01820546 & \qquad lorikeet\\
\qquad n02097658 & \qquad silky terrier & \qquad n02105251 & \qquad briard & \qquad n01843383 & \qquad toucan\\
\qquad n02098286 & \qquad West Highland white  & \qquad n02105855 & \qquad Shetland sheepdog & \qquad n02002724 & \qquad black stork\\
\qquad n02099267 & \qquad flat-coated retrieve & \qquad n02106166 & \qquad Border collie & \qquad n02012849 & \qquad crane\\
\qquad n02099601 & \qquad golden retriever & \qquad n02107142 & \qquad Doberman & \qquad n02027492 & \qquad red-backed sandpiper\\
\qquad n02099849 & \qquad Chesapeake Bay retri & \qquad n02107574 & \qquad Greater Swiss Mounta & \qquad n02058221 & \qquad albatross\\
\qquad n02100583 & \qquad vizsla & \qquad n02108089 & \qquad boxer & \it n02159955 & \it insect\\
\qquad n02100877 & \qquad Irish setter & \qquad n02108551 & \qquad Tibetan mastiff & \qquad n02168699 & \qquad long-horned beetle\\
\qquad n02101388 & \qquad Brittany spaniel & \qquad n02109961 & \qquad Eskimo dog & \qquad n02206856 & \qquad bee\\
\qquad n02102040 & \qquad English springer & \qquad n02110185 & \qquad Siberian husky & \qquad n02236044 & \qquad mantis\\
\qquad n02102318 & \qquad cocker spaniel & \qquad n02110958 & \qquad pug & \qquad n02276258 & \qquad admiral\\
\qquad n02102973 & \qquad Irish water spaniel & \qquad n02111277 & \qquad Newfoundland & \it n03405725 & \it furniture\\
\qquad n02104365 & \qquad schipperke & \qquad n02112137 & \qquad chow & \qquad n03016953 & \qquad chiffonier\\
\qquad n02105162 & \qquad malinois & \qquad n02112706 & \qquad Brabancon griffon & \qquad n03290653 & \qquad entertainment center\\
\qquad n02105412 & \qquad kelpie & \qquad n02113712 & \qquad miniature poodle & \qquad n04099969 & \qquad rocking chair\\
\qquad n02105641 & \qquad Old English sheepdog & \qquad n02113978 & \qquad Mexican hairless & \it n02512053 & \it fish\\
\qquad n02106030 & \qquad collie & \it n01503061 & \it bird & \qquad n01491361 & \qquad tiger shark\\
\qquad n02106382 & \qquad Bouvier des Flandres & \qquad n01514859 & \qquad hen & \qquad n02536864 & \qquad coho\\
\qquad n02106662 & \qquad German shepherd & \qquad n01532829 & \qquad house finch & \qquad n02655020 & \qquad puffer\\
\qquad n02107312 & \qquad miniature pinscher & \qquad n01537544 & \qquad indigo bunting & \it n02484322 & \it monkey\\
\qquad n02107683 & \qquad Bernese mountain dog & \qquad n01582220 & \qquad magpie & \qquad n02487347 & \qquad macaque\\
\qquad n02108000 & \qquad EntleBucher & \qquad n01601694 & \qquad water ouzel & \qquad n02492660 & \qquad howler monkey\\
\qquad n02108422 & \qquad bull mastiff & \qquad n01622779 & \qquad great grey owl & \it n02958343 & \it car\\
\qquad n02108915 & \qquad French bulldog & \qquad n01818515 & \qquad macaw & \qquad n03100240 & \qquad convertible\\
\qquad n02109525 & \qquad Saint Bernard & \qquad n01828970 & \qquad bee eater & \qquad n04285008 & \qquad sports car\\
\qquad n02110063 & \qquad malamute & \qquad n01833805 & \qquad hummingbird & \it n02120997 & \it feline\\
\qquad n02110341 & \qquad dalmatian & \qquad n01855032 & \qquad red-breasted mergans & \qquad n02128757 & \qquad snow leopard\\
\qquad n02110806 & \qquad basenji & \qquad n01860187 & \qquad black swan & \it n04490091 & \it truck\\
\qquad n02111129 & \qquad Leonberg & \qquad n02007558 & \qquad flamingo & \qquad n03930630 & \qquad pickup\\
\qquad n02111500 & \qquad Great Pyrenees & \qquad n02009912 & \qquad American egret & \it n13134947 & \it fruit\\
\qquad n02112018 & \qquad Pomeranian & \qquad n02017213 & \qquad European gallinule & \qquad n12267677 & \qquad acorn\\
\qquad n02112350 & \qquad keeshond & \qquad n02018795 & \qquad bustard & \it n12992868 & \it fungus\\
\qquad n02113023 & \qquad Pembroke & \qquad n02033041 & \qquad dowitcher & \qquad n13040303 & \qquad stinkhorn\\
\qquad n02113624 & \qquad toy poodle & \qquad n02051845 & \qquad pelican & \it n02858304 & \it boat\\
\qquad n02113799 & \qquad standard poodle & \it n02159955 & \it insect & \qquad n03662601 & \qquad lifeboat\\
\it n01503061 & \it bird & \qquad n02165456 & \qquad ladybug & \it n03082979 & \it computer\\
\qquad n01514668 & \qquad cock & \qquad n02172182 & \qquad dung beetle & \qquad n03832673 & \qquad notebook\\
\qquad n01518878 & \qquad ostrich & \qquad n02177972 & \qquad weevil & \it n01661091 & \it reptile\\
\qquad n01531178 & \qquad goldfinch & \qquad n02226429 & \qquad grasshopper & \qquad n01664065 & \qquad loggerhead\\
\qquad n01534433 & \qquad junco & \qquad n02231487 & \qquad walking stick & \qquad n01665541 & \qquad leatherback turtle\\
\qquad n01558993 & \qquad robin & \qquad n02259212 & \qquad leafhopper & \qquad n01667114 & \qquad mud turtle\\
\qquad n01580077 & \qquad jay & \qquad n02268443 & \qquad dragonfly & \qquad n01667778 & \qquad terrapin\\
\qquad n01592084 & \qquad chickadee & \qquad n02279972 & \qquad monarch & \qquad n01669191 & \qquad box turtle\\
\qquad n01608432 & \qquad kite & \qquad n02281406 & \qquad sulphur butterfly & \qquad n01675722 & \qquad banded gecko\\
\qquad n01616318 & \qquad vulture & \it n03405725 & \it furniture & \qquad n01677366 & \qquad common iguana\\
\qquad n01817953 & \qquad African grey & \qquad n02804414 & \qquad bassinet & \qquad n01682714 & \qquad American chameleon\\
\qquad n01819313 & \qquad sulphur-crested cock & \qquad n03125729 & \qquad cradle & \qquad n01685808 & \qquad whiptail\\
\qquad n01824575 & \qquad coucal & \qquad n03179701 & \qquad desk & \qquad n01687978 & \qquad agama\\
\qquad n01829413 & \qquad hornbill & \qquad n03376595 & \qquad folding chair & \qquad n01688243 & \qquad frilled lizard\\
\qquad n01843065 & \qquad jacamar & \qquad n03742115 & \qquad medicine chest & \qquad n01689811 & \qquad alligator lizard\\
\qquad n01847000 & \qquad drake & \qquad n04380533 & \qquad table lamp & \qquad n01692333 & \qquad Gila monster\\
\qquad n01855672 & \qquad goose & \qquad n04447861 & \qquad toilet seat & \qquad n01693334 & \qquad green lizard\\
\qquad n02002556 & \qquad white stork & \it n02512053 & \it fish & \qquad n01694178 & \qquad African chameleon\\
\qquad n02006656 & \qquad spoonbill & \qquad n01443537 & \qquad goldfish & \qquad n01695060 & \qquad Komodo dragon\\
\qquad n02009229 & \qquad little blue heron & \qquad n01496331 & \qquad electric ray & \qquad n01697457 & \qquad African crocodile\\
\qquad n02011460 & \qquad bittern & \qquad n02514041 & \qquad barracouta & \qquad n01698640 & \qquad American alligator\\
\qquad n02013706 & \qquad limpkin & \qquad n02607072 & \qquad anemone fish & \qquad n01704323 & \qquad triceratops\\
\qquad n02018207 & \qquad American coot & \qquad n02641379 & \qquad gar & \qquad n01728572 & \qquad thunder snake\\
\qquad n02025239 & \qquad ruddy turnstone & \it n02484322 & \it monkey & \qquad n01728920 & \qquad ringneck snake\\
\qquad n02028035 & \qquad redshank & \qquad n02486261 & \qquad patas & \qquad n01729322 & \qquad hognose snake\\
\qquad n02037110 & \qquad oystercatcher & \qquad n02488702 & \qquad colobus & \qquad n01729977 & \qquad green snake\\
\qquad n02056570 & \qquad king penguin & \qquad n02490219 & \qquad marmoset & \qquad n01734418 & \qquad king snake\\
\it n02159955 & \it insect & \qquad n02493793 & \qquad spider monkey & \qquad n01735189 & \qquad garter snake\\
\qquad n02165105 & \qquad tiger beetle & \it n02958343 & \it car & \qquad n01737021 & \qquad water snake\\
\qquad n02167151 & \qquad ground beetle & \qquad n02814533 & \qquad beach wagon & \qquad n01739381 & \qquad vine snake\\
\qquad n02169497 & \qquad leaf beetle & \qquad n03670208 & \qquad limousine & \qquad n01740131 & \qquad night snake\\
\qquad n02174001 & \qquad rhinoceros beetle & \qquad n03777568 & \qquad Model T & \qquad n01742172 & \qquad boa constrictor\\
\qquad n02190166 & \qquad fly & \it n02120997 & \it feline & \qquad n01744401 & \qquad rock python\\
\qquad n02219486 & \qquad ant & \qquad n02127052 & \qquad lynx & \qquad n01748264 & \qquad Indian cobra\\
\qquad n02229544 & \qquad cricket & \qquad n02129165 & \qquad lion & \qquad n01749939 & \qquad green mamba\\
\qquad n02233338 & \qquad cockroach & \qquad n02130308 & \qquad cheetah & \qquad n01751748 & \qquad sea snake\\
\qquad n02256656 & \qquad cicada & \it n04490091 & \it truck & \qquad n01753488 & \qquad horned viper\\
\qquad n02264363 & \qquad lacewing & \qquad n03417042 & \qquad garbage truck & \qquad n01755581 & \qquad diamondback\\
\qquad n02268853 & \qquad damselfly & \qquad n04461696 & \qquad tow truck & \qquad n01756291 & \qquad sidewinder\\
\qquad n02277742 & \qquad ringlet & \it n13134947 & \it fruit & \it n03051540 & \it clothing\\
\qquad n02280649 & \qquad cabbage butterfly & \qquad n11879895 & \qquad rapeseed & \qquad n02667093 & \qquad abaya\\
\qquad n02281787 & \qquad lycaenid & \qquad n12768682 & \qquad buckeye & \qquad n02669723 & \qquad academic gown\\
\it n03405725 & \it furniture & \it n12992868 & \it fungus & \qquad n02730930 & \qquad apron\\
\qquad n02791124 & \qquad barber chair & \qquad n12998815 & \qquad agaric & \qquad n02807133 & \qquad bathing cap\\
\qquad n02870880 & \qquad bookcase & \qquad n13052670 & \qquad hen-of-the-woods & \qquad n02817516 & \qquad bearskin\\
\qquad n03018349 & \qquad china cabinet & \it n02858304 & \it boat & \qquad n02837789 & \qquad bikini\\
\qquad n03131574 & \qquad crib & \qquad n03344393 & \qquad fireboat & \qquad n02865351 & \qquad bolo tie\\
\qquad n03201208 & \qquad dining table & \qquad n04612504 & \qquad yawl & \qquad n02869837 & \qquad bonnet\\
\qquad n03337140 & \qquad file & \it n03082979 & \it computer & \qquad n02883205 & \qquad bow tie\\
\qquad n03388549 & \qquad four-poster & \qquad n03485407 & \qquad hand-held computer & \qquad n02892767 & \qquad brassiere\\
\qquad n03891251 & \qquad park bench & \qquad n06359193 & \qquad web site & \qquad n02963159 & \qquad cardigan\\
\qquad n04344873 & \qquad studio couch &  &  & \qquad n03026506 & \qquad Christmas stocking\\
\qquad n04429376 & \qquad throne &  &  & \qquad n03124170 & \qquad cowboy hat\\
\qquad n04550184 & \qquad wardrobe &  &  & \qquad n03127747 & \qquad crash helmet\\
\it n02512053 & \it fish &  &  & \qquad n03188531 & \qquad diaper\\
\qquad n01440764 & \qquad tench &  &  & \qquad n03325584 & \qquad feather boa\\
\qquad n01484850 & \qquad great white shark &  &  & \qquad n03379051 & \qquad football helmet\\
\qquad n01494475 & \qquad hammerhead &  &  & \qquad n03404251 & \qquad fur coat\\
\qquad n01498041 & \qquad stingray &  &  & \qquad n03450230 & \qquad gown\\
\qquad n02526121 & \qquad eel &  &  & \qquad n03534580 & \qquad hoopskirt\\
\qquad n02606052 & \qquad rock beauty &  &  & \qquad n03594734 & \qquad jean\\
\qquad n02640242 & \qquad sturgeon &  &  & \qquad n03595614 & \qquad jersey\\
\qquad n02643566 & \qquad lionfish &  &  & \qquad n03617480 & \qquad kimono\\
\it n02484322 & \it monkey &  &  & \qquad n03623198 & \qquad knee pad\\
\qquad n02484975 & \qquad guenon &  &  & \qquad n03630383 & \qquad lab coat\\
\qquad n02486410 & \qquad baboon &  &  & \qquad n03710637 & \qquad maillot\\
\qquad n02488291 & \qquad langur &  &  & \qquad n03710721 & \qquad maillot\\
\qquad n02489166 & \qquad proboscis monkey &  &  & \qquad n03724870 & \qquad mask\\
\qquad n02492035 & \qquad capuchin &  &  & \qquad n03763968 & \qquad military uniform\\
\qquad n02493509 & \qquad titi &  &  & \qquad n03770439 & \qquad miniskirt\\
\qquad n02494079 & \qquad squirrel monkey &  &  & \qquad n03775071 & \qquad mitten\\
\it n02958343 & \it car &  &  & \qquad n03787032 & \qquad mortarboard\\
\qquad n02701002 & \qquad ambulance &  &  & \qquad n03866082 & \qquad overskirt\\
\qquad n02930766 & \qquad cab &  &  & \qquad n03877472 & \qquad pajama\\
\qquad n03594945 & \qquad jeep &  &  & \qquad n03980874 & \qquad poncho\\
\qquad n03770679 & \qquad minivan &  &  & \qquad n04136333 & \qquad sarong\\
\qquad n04037443 & \qquad racer &  &  & \qquad n04162706 & \qquad seat belt\\
\it n02120997 & \it feline &  &  & \qquad n04209133 & \qquad shower cap\\
\qquad n02125311 & \qquad cougar &  &  & \qquad n04254777 & \qquad sock\\
\qquad n02128385 & \qquad leopard &  &  & \qquad n04259630 & \qquad sombrero\\
\qquad n02128925 & \qquad jaguar &  &  & \qquad n04325704 & \qquad stole\\
\qquad n02129604 & \qquad tiger &  &  & \qquad n04350905 & \qquad suit\\
\it n04490091 & \it truck &  &  & \qquad n04370456 & \qquad sweatshirt\\
\qquad n03345487 & \qquad fire engine &  &  & \qquad n04371430 & \qquad swimming trunks\\
\qquad n03796401 & \qquad moving van &  &  & \qquad n04479046 & \qquad trench coat\\
\qquad n03977966 & \qquad police van &  &  & \qquad n04532106 & \qquad vestment\\
\qquad n04467665 & \qquad trailer truck &  &  & \qquad n04584207 & \qquad wig\\
\it n13134947 & \it fruit &  &  & \qquad n04591157 & \qquad Windsor tie\\
\qquad n07742313 & \qquad Granny Smith &  &  & \it n02370806 & \it ungulate\\
\qquad n12144580 & \qquad corn &  &  & \qquad n02389026 & \qquad sorrel\\
\qquad n12620546 & \qquad hip &  &  & \qquad n02391049 & \qquad zebra\\
\qquad n13133613 & \qquad ear &  &  & \qquad n02395406 & \qquad hog\\
\it n12992868 & \it fungus &  &  & \qquad n02396427 & \qquad wild boar\\
\qquad n12985857 & \qquad coral fungus &  &  & \qquad n02397096 & \qquad warthog\\
\qquad n13037406 & \qquad gyromitra &  &  & \qquad n02398521 & \qquad hippopotamus\\
\qquad n13044778 & \qquad earthstar &  &  & \qquad n02403003 & \qquad ox\\
\qquad n13054560 & \qquad bolete &  &  & \qquad n02408429 & \qquad water buffalo\\
\it n02858304 & \it boat &  &  & \qquad n02410509 & \qquad bison\\
\qquad n02951358 & \qquad canoe &  &  & \qquad n02412080 & \qquad ram\\
\qquad n03447447 & \qquad gondola &  &  & \qquad n02415577 & \qquad bighorn\\
\qquad n04273569 & \qquad speedboat &  &  & \qquad n02417914 & \qquad ibex\\
\it n03082979 & \it computer &  &  & \qquad n02422106 & \qquad hartebeest\\
\qquad n03180011 & \qquad desktop computer &  &  & \qquad n02422699 & \qquad impala\\
\qquad n03642806 & \qquad laptop &  &  & \qquad n02423022 & \qquad gazelle\\
\qquad n04238763 & \qquad slide rule &  &  & \qquad n02437312 & \qquad Arabian camel\\
 &  &  &  & \qquad n02437616 & \qquad llama\\
 &  &  &  & \it n07707451 & \it vegetable\\
 &  &  &  & \qquad n07714571 & \qquad head cabbage\\
 &  &  &  & \qquad n07714990 & \qquad broccoli\\
 &  &  &  & \qquad n07715103 & \qquad cauliflower\\
 &  &  &  & \qquad n07716358 & \qquad zucchini\\
 &  &  &  & \qquad n07716906 & \qquad spaghetti squash\\
 &  &  &  & \qquad n07717410 & \qquad acorn squash\\
 &  &  &  & \qquad n07717556 & \qquad butternut squash\\
 &  &  &  & \qquad n07718472 & \qquad cucumber\\
 &  &  &  & \qquad n07718747 & \qquad artichoke\\
 &  &  &  & \qquad n07720875 & \qquad bell pepper\\
 &  &  &  & \qquad n07730033 & \qquad cardoon\\
 &  &  &  & \qquad n07734744 & \qquad mushroom\\
 &  &  &  & \it n02686568 & \it aircraft\\
 &  &  &  & \qquad n02690373 & \qquad airliner\\
 &  &  &  & \qquad n02692877 & \qquad airship\\
 &  &  &  & \qquad n02782093 & \qquad balloon\\

\end{longtable}

\end{document}